\documentclass[onecolumn]{IEEEtran}
\IEEEoverridecommandlockouts  
\usepackage[utf8]{inputenc}
\usepackage{graphics} 
\usepackage{epsfig} 
\usepackage{mathptmx,xcolor} 
\usepackage{times} 
\usepackage{amsmath} 
\usepackage{amssymb}  
\usepackage{mathrsfs}
\usepackage{graphicx}
\usepackage{caption}
\usepackage{subcaption}
\usepackage{algorithmic}
\usepackage{algorithm}
\usepackage{hyperref}
\usepackage{graphicx}
\usepackage{bbm}
\usepackage{enumerate}
\usepackage{theorem}
\usepackage{makecell}
\input{mysymbol.sty}
\usepackage{needspace}

\newcommand{\closure}[2][3]{{}\mkern#1mu\overline{\mkern-#1mu#2}}
\DeclareMathAlphabet{\mathcal}{OMS}{cmsy}{m}{n}
\newcommand{\algorithmicbreak}{\textbf{break}}
\newcommand{\BREAK}{\STATE \algorithmicbreak}

\newtheorem{assumption}{\hspace{0pt}\bf Assumption}
\newtheorem{lemma}{\hspace{0pt}\bf Lemma}
\newtheorem{theorem}{\hspace{0pt}\bf Theorem}
\newtheorem{proposition}{\hspace{0pt}\bf Proposition}

\newtheorem{remark}{\hspace{0pt}\bf Remark}
\makeatletter

\title{\LARGE \bf  Nonparametric Stochastic Compositional Gradient Descent\\ for  Q-Learning in Continuous Markov Decision Problems}
\author{ Alec Koppel*$^\dagger$, Ekaterina Tolstaya*$^\star$, Ethan Stump$^\dagger$, Alejandro Ribeiro$^\star$
\thanks{* Equally contributing authors. Supported by NSF DGE-1321851, ARL DCIST CRA W911NF-17-2-0181, Intel DevCloud and Intel Science and Technology Center for Wireless Autonomous Systems (ISTC-WAS). $^\star$Department of ESE, University of Pennsylvania. Email: \{eig, aribeiro\}@seas.upenn.edu. $^\dagger$Computational and Information Sciences Directorate, U.S. Army Research Laboratory. Email: \{alec.e.koppel.civ, ethan.a.stump2.civ\}@mail.mil.  
 Part of the results in this paper appeared in \cite{tolstaya2018}. This paper expands the algorithm's applicability, elaborates on implementation issues, includes proofs of theoretical results, and a more thorough numerical evaluation.} 
}
\begin{document}
\maketitle
\thispagestyle{empty}
\pagestyle{empty}
\begin{abstract}
We consider Markov Decision Problems defined over continuous state and action spaces, where an autonomous agent seeks to learn a map from its states to actions so as to maximize its long-term discounted accumulation of rewards. We address this problem by considering Bellman's optimality equation defined over action-value functions, which we reformulate into a nested non-convex stochastic optimization problem defined over a Reproducing Kernel Hilbert Space (RKHS). We develop a functional generalization of stochastic quasi-gradient method to solve it, which, owing to the structure of the RKHS, admits a parameterization in terms of scalar weights and past state-action pairs which grows proportionately with the algorithm iteration index. To ameliorate this complexity explosion, we apply Kernel Orthogonal Matching Pursuit to the sequence of kernel weights and dictionaries, which yields a controllable error in the descent direction of the underlying optimization method. We prove that the resulting algorithm, called KQ-Learning, converges with probability $1$ to a stationary point of this problem, yielding a fixed point of the Bellman optimality operator under the hypothesis that it belongs to the RKHS. Under constant learning rates, we further obtain convergence to a small Bellman error that depends on the chosen learning rates. Numerical evaluation on the Continuous Mountain Car and Inverted Pendulum tasks yields convergent parsimonious learned action-value functions,  policies that are competitive with the state of the art, and exhibit reliable, reproducible learning behavior.
\end{abstract}

\section{Introduction}\label{sec:intro}

Markov Decision Problems offer a flexible framework to address sequential decision making tasks under uncertainty \cite{bellman1954theory}, and have gained broad interest in robotics \cite{kober2013reinforcement}, control \cite{bertsekas1978stochastic}, finance \cite{rasonyi2005utility}, and artificial intelligence \cite{sutton2018reinforcement}. Despite this surge of interest, few works in reinforcement learning address the computational difficulties associated with continuous state and action spaces in a principled way that guarantees convergence. The goal of this work is to develop new reinforcement learning tools for continuous problems which are provably stable and whose complexity is at-worst moderate.

In the development of stochastic methods for reinforcement learning, one may attempt to estimate the transition density of the Markov Decision Process (MDP) (model-based \cite{mitkovska2014methodology}), perform gradient descent on the value function with respect to the policy (direct policy search \cite{sutton2000policy}), and pursue value function based (model-free \cite{sutton1988learning,Watkins1989}) methods which exploit structural properties of the setting to derive fixed point problems called \emph{Bellman equations}. We adopt the latter approach in this work \cite{powell2011review}, motivated by the fact that an action-value function tells us both how to find a policy and how to evaluate it in terms of the performance metric we have defined, and that a value function encapsulates structural properties of the relationship between states, actions, and rewards.

It is well-known that approaches featuring the ``deadly triad'' \cite{sutton2018reinforcement} of function approximation, bootstrapping (e.g. temporal-difference learning), and off-policy training are in danger of divergence, and the most well-understood techniques for ensuring convergence in a stochastic gradient descent context are those based on Gradient Temporal Difference (GTD) \cite{sutton2009convergent}. Though the final algorithm looks similar, our approach could be considered as an alternative formulation and analysis of the GTD family of algorithms centered on a flexible RKHS representation that lets us address problems with nonlinear, continuous state and action spaces in a natural way.

To understand our proposed approach, consider the fixed point problem defined by Bellman's optimality equation \cite{Bellman:1957}. When the state and action spaces are finite and small enough that expectations are computable, fixed point iterations may be used. When this fails to hold, stochastic fixed point methods, namely, $Q$-learning \cite{Watkins1989}, may be used, whose convergence may be addressed with asynchronous stochastic approximation theory \cite{tsitsiklis1994asynchronous,jaakkola1994convergence}. This approach is only valid when the action-value (or $Q$) function may be represented as a matrix. However, when the state and action spaces are infinite, this is no longer true, and the $Q$-function instead belongs to a generic function space. 

In particular, to solve the fixed point problem defined by Bellman's optimality equation when spaces are continuous, one must surmount the fact that it is defined for infinitely many unknowns, one example of Bellman's curse of dimensionality \cite{Bellman:1957}. Efforts to sidestep this issue assume that the $Q$-function admits a finite parameterization, such as a linear \cite{melo2008analysis,sutton2009convergent} or nonlinear \cite{bhatnagar2009convergent} basis expansion, is defined by a neural network \cite{mnih2013playing}, or that it belongs to a reproducing kernel Hilbert Space (RKHS) \cite{kimeldorf1971some,scholkopfgeneralized}. In this work, we adopt the later nonparametric approach, motivated by the fact that combining fixed point iterations with different parameterizations may cause divergence \cite{Baird95residualalgorithms,tsitsiklis1997analysis}, and in general the $Q$-function parameterization must be tied to the stochastic update to ensure the convergence of both the function sequence and its parameterization \cite{jong2007model}.

Our main result is a memory-efficient, non-parametric, stochastic method that converges to a fixed point of the Bellman optimality operator almost surely when it belongs to a RKHS. We obtain this result by reformulating the Bellman optimality equation as a nested stochastic program (Section \ref{sec:prob}), a topic investigated in operations research \cite{shapiro2014lectures} and probability \cite{Korostelev,konda2004convergence}. These problems have been addressed in finite settings with stochastic \emph{quasi-gradient} (SQG) methods \cite{ermoliev1983stochastic} which use two time-scale stochastic approximation to mitigate the fact that the objective's stochastic gradient not available due to its dependence on a \emph{second expectation}, which is referred to as the double sampling problem in \cite{sutton2009convergent}. 

Here, we use a non-parametric generalization of SQG for $Q$-learning in infinite MDPs (Section \ref{sec:alg}), motivated by its success for policy evaluation in finite \cite{sutton2009convergent,bhatnagar2009convergent} and infinite MDPs \cite{pkgtd}. However, a function in a RKHS has comparable complexity to the {number of training samples processed}, which is in general infinite, an issue is often ignored in kernel methods for Markov decision problems \cite{ormoneit2002kernel,grunewalder2012modelling,JMLR:v17:13-016,dai2016learning}. We address this bottleneck (the curse of kernelization) by requiring memory efficiency in both the function sample path and in its limit through the use of sparse projections which are constructed greedily via matching pursuit \cite{mallat1993matching,lever2016compressed}, akin to \cite{polk,pkgtd}. Greedy compression is appropriate since (a) kernel matrices induced by arbitrary data streams will likely become ill-conditioned and hence violate assumptions required by convex methods \cite{candes2008restricted}, and (b) parsimony is more important than exact recovery as the SQG iterates are not the target signal but rather a stepping stone to Bellman fixed point.
Rather than unsupervised forgetting \cite{kivinen2002online}, we tie the projection-induced error to guarantee stochastic descent \cite{polk}, only keeping dictionary points needed for convergence.
 
As a result, we conduct functional SQG descent via sparse projections of the SQG. This maintains a {moderate-complexity sample path exactly towards $Q^*$}, which may be made arbitrarily close to a Bellman fixed point by decreasing the regularizer.
In contrast to the convex structure in \cite{pkgtd}, the Bellman optimality equation induces a non-convex cost functional, which requires us to generalize the relationship between SQG for non-convex objectives and coupled supermartingales in \cite{wang2017stochastic} to RKHSs. In doing so, we establish that the sparse projected SQG sequence converges almost surely (Theorem \ref{main_theorem}) to the Bellman fixed point with decreasing learning rates (Section \ref{sec:convergence}) and to a small Bellman error whose magnitude depends on the learning rates when learning rates are held constant (Theorem \ref{main_theorem2}). Use of constant learning rates allows us to further guarantee that the memory of the learned $Q$ function remains under control. Moreover, on Continuous Mountain Car \cite{Moore90efficientmemory} and the Inverted Pendulum \cite{yoshida1999swing}, we observe that our learned action-value function attains a favorable trade-off between memory efficiency and Bellman error, which then yields a policy whose performance is competitive with the state of the art in terms of episode average reward accumulation.

\section{Markov Decision Processes}\label{sec:prob}
We model an autonomous agent in a continuous space as a Markov Decision Process (MDP) with continuous states $\bbs\in\ccalS \subset \mathbb{R}^p$ and actions $\bba\in\mathcal{A} \subset \mathbb{R}^q$. When in state $\bbs$ and taking action $\bba$, a random transition to state $s'$ occurs according to the conditional probability density $\mbP(\bbs' | \bbs, \bba)$. 
After the agent transitions to a particular $\bbs'$ from $\bbs$, the MDP assigns an instantaneous reward $r(\bbs,\bba,\bbs')$, where the reward function is a map $r: \ccalS \times \mathcal{A} \times \ccalS \rightarrow \mathbb{R}$.  

In Markov Decision problems, the goal is to find the action sequence $\{ \bba_t \}_{t=0}^\infty$ so as to maximize the infinite horizon accumulation of rewards, i.e.,  the value function: $V(\bbs,\{ \bba_t \}_{t=0}^\infty) := \mathbb{E}_{\bbs'} \lbrack \sum_{t=0}^\infty \gamma^t r(\bbs_t,\bba_t,\bbs'_t) \mid \bbs_0 = \bbs, \{ \bba_t \}_{t=0}^\infty  \rbrack$. The action-value function $Q(\bbs,\bba)$ is the conditional mean of the value function given the initial action $\bba_0=\bba$:
\begin{align} \label{eq:qdef}
Q(\!\bbs,\bba,\!\!\{\!\bba_t\! \}_{t=1}^\infty\!) 
\!\!:= \! \mathbb{E}_{\bbs'}  \!\!\!\left[\!  \sum_{t=0}^\infty \!\gamma^t \! r(\bbs_t,\bba_t,\bbs'_t) \!\!   \mid \!\bbs_0\!\! =\! \bbs, \bba_0\! = \!\bba, \! \{ \!\bba_t \!\}_{t=1}^\infty  \!\!\right] 
\end{align}
 We define $Q^*(\bbs,\bba)$ as the maximum of \eqref{eq:qdef} with respect to the action sequence. The reason for defining action-value functions is that the optimal  $Q^*$ may be used to compute the optimal policy $\pi^*$ as
\begin{equation} \label{eq:pistar} 
\pi^*(\bbs) = \argmax_\bba Q^*(\bbs,\bba) \; .
\end{equation}
where a policy is a map from states to actions: $\pi$: $\ccalS \to \mathcal{A}$.
Thus, finding $Q^*$ solves the MDP. Value-function based approaches to MDPs reformulate \eqref{eq:pistar} by shifting the index of the summand in \eqref{eq:qdef} by one, use the time invariance of the Markov transition kernel, and the homogeneity of the summand, to derive the Bellman optimality equation:
\begin{equation} \label{eq:qstar} 
   Q^{*}(\bbs,\bba) =   \mbE_{\bbs'} \Big[ 
                        r(\bbs,\bba,\bbs') + \gamma\max_{\bba'}Q^{*}(\bbs',\bba')  
                        \given \bbs, \bba \Big] .
\end{equation}
where the expectation is taken with respect to the conditional distribution $\mbP(d\bbs' \mid \bbs, \bba)$ of the state $\bbs'$ given the state action pair $(\bbs,\bba)$. The right-hand side of Equation \eqref{eq:qstar} defines the Bellman optimality operator $\mathscr{B}^*$: $\ccalB(\ccalS \times \mathcal{A}) \to \ccalB(\ccalS \times \mathcal{A})$ over $\ccalB(\ccalS \times \mathcal{A})$, the space of bounded continuous action-value functions Q: $\ccalB(\ccalS \times \mathcal{A}) \to \mathbb{R}$: 
\begin{align} \label{eq:bellop} 
\!\!\!\!\!\!(\!\mathscr{B}^* \!Q)(\bbs,\bba)\! :=\! \!\!\int_{\ccalS} \! \lbrack r(\!\bbs,\!\bba,\!\bbs') \!+\!  \gamma  \max_{\bba'} Q(\bbs'\!,\bba') \rbrack  \mbP(\!d\bbs' \!\!\mid \bbs, \bba\!)\!\! \; .
\end{align}
\cite{bertsekas1978stochastic} [Proposition 5.2] establishes that the fixed point of \eqref{eq:bellop} is the optimal action-value function $Q^*$. Thus, to solve the MDP, we seek to compute the fixed point of \eqref{eq:bellop} for all $(\bbs,\bba) \in \ccalS\times \mathcal{A}$.

%
\smallskip\noindent\textbf{Compositional Stochastic Optimization.} The functional fixed point equation in \eqref{eq:qstar} has to be simultaneously satisfied for all state action pairs $(\bbs,\bba)$. Alternatively, we can integrate \eqref{eq:qstar} over an arbitrary distribution that is dense around any pair $(\bbs,\bba)$ to write a nested stochastic optimization problem \cite{wang2017stochastic, polk,pkgtd}. To do so, begin by defining the function 
\begin{align}  \label{eq:f} 
   f(Q; \bbs\!,\bba)\!= \mbE_{\bbs'}\!\Big[ 
                        r(\bbs\!,\bba\!,\bbs')+ \gamma\max_{\bba'}Q(\bbs'\!,\bba')  
                        - Q(\bbs,\bba)
                        \given \bbs\!,\bba \Big]\!,
\end{align}
and consider an arbitrary everywhere dense distribution $\mbP(d\bbs, d\bba)$ over pairs $(\bbs,\bba)$ to define the functional 
\begin{align}  \label{eq:cost_functional} 
  L(Q) = \frac{1}{2}\mbE_{\bbs,\bba} \Big[ f^2(Q; \bbs,\bba) \Big].
\end{align}
Comparing \eqref{eq:f} with \eqref{eq:qstar} permits concluding that $\bbQ^*$ is the unique function that makes $f(Q; \bbs,\bba)=0$ for all $(\bbs,\bba)$. It then follows that $\bbQ^*$ is the only function that makes the functional in \eqref{eq:cost_functional} take the value $L(Q)=0$. Since this functional is also nonnegative, we can write the optimal $Q$ function as
\begin{align}  \label{eq:main_prob} 
   Q^{*} = \argmin_{Q\in \ccalB(\ccalS\times\mathcal{A})} L(Q) \; .
\end{align}
%
Computation of the optimal policy is thus equivalent to solving the optimization problem in \eqref{eq:main_prob}. This requires a difficult search over all bounded continuous functions $\ccalB(\ccalS\times\mathcal{A})$. We reduce this difficulty through a hypothesis on the function class.

\smallskip\noindent\textbf{Reproducing Kernel Hilbert Spaces} 
We propose restricting $\mathcal{B}(\ccalS\times\mathcal{A})$ to be a Hilbert space $\ccalH$ equipped with a unique reproducing kernel, an inner product-like map $\kappa:(\ccalS\times\mathcal{A})\times(\ccalS\times\mathcal{A}) \rightarrow \mathbb{R}$ such that
\begin{align} \label{eq:repprop}
\!\!\!\!\! \text{(i)} \ \langle f\!, \! \kappa((\bbs,\bba),\! \cdot \!) \rangle_\ccalH \!\! = \!\! f(\bbs,\bba\!) ,\quad
\text{(ii)} \ \ccalH \!=\! \closure{ \text{span} \lbrace \!\kappa(\!(\bbs,\bba\!),\!\cdot\!) \rbrace }
\end{align}
In \eqref{eq:repprop}, property (i) is called the reproducing property. Replacing $f$ by $\kappa((\bbs',\bba'),\cdot)$ in \eqref{eq:repprop} (i) yields the expression $\langle \kappa((\bbs',\bba'),\cdot)\!, \! \kappa((\bbs,\bba),\! \cdot \!) \!\rangle_\ccalH = \kappa((\bbs',\bba'),(\bbs,\bba))$, the origin of the term “reproducing kernel.” Moreover, property \eqref{eq:repprop} (ii) states that functions $f \in \ccalH$ admit a basis expansion in terms of kernel evaluations \eqref{eq:qrkhs}. Function spaces of this type are referred to as reproducing kernel Hilbert spaces (RKHSs).

We may apply the Representer Theorem to transform the functional problem into a parametric one 
\cite{norkin2009stochastic,argyriou2009there}. 
In the Reproducing Kernel Hilbert Space (RKHS), the optimal $Q$ function takes the following form
\begin{equation}  \label{eq:qrkhs} 
Q(\bbs,\bba) = \sum_{n=1}^N w_n \kappa((\bbs_n,\bba_n),(\bbs,\bba)) 
\end{equation}
where $(\bbs_n,\bba_n)$ is a sample of state-action pairs $(\bbs, \bba)$. $Q \in \ccalH$ is an expansion of kernel evaluations only at observed samples.

One complication of the restriction  $\mathcal{B}(\ccalS\times\mathcal{A})$ to the RKHS $\ccalH$ is that this setting requires the cost to be differentiable with Lipschitz gradients, but the definition of $ L(Q)$ [cf. \eqref{eq:cost_functional}] defined by Bellman's equation \eqref{eq:bellop} is non-differentiable due to the presence of the maximization over the $Q$ function. This issue may be addressed by either operating with approximate smoothed gradients of a non-differentiable function \cite{nesterov2005smooth} or by approximating the non-smooth cost by a smooth one. We adopt the latter approach by replacing the $\max_\bba Q(\bbs, \bba')$ term in \eqref{eq:cost_functional} by the softmax over continuous range $\ccalA$, i.e.
\begin{equation}\label{eq:softmax}
\text{softmax}_{\bba' \in \ccalA} Q(\bbs', \bba') = \frac{1}{\eta}\log\int_{\bba' \in \ccalA} e^{\eta Q(\bbs',\bba')} d\bba'
\end{equation}
and define the $\eta$-smoothed cost $L(Q)$ as the one where the softmax \eqref{eq:softmax} in lieu of the hard maximum in \eqref{eq:cost_functional}. Subsequently, we restrict focus to smoothed cost $L(Q)$.

In this work, we restrict the kernel used to be in the family of universal kernels, such as a Gaussian Gaussian Radial Basis Function(RBF) kernel with constant diagonal covariance $\Sigma$, 
\begin{align}  \label{eq:rbfdef}
\!\!\kappa(\!(\!\mathbf{s},\mathbf{a}),\!(\mathbf{s}'\!\!,\mathbf{a}')\!) \! =\!  \exp \!\lbrace \! - \frac{1}{2} \!(\!(\mathbf{s},\mathbf{a}\!)\!-\!(\!\mathbf{s}'\!\!,\mathbf{a}')\!)\Sigma (\!(\!\mathbf{s},\mathbf{a}\!)\!-\!(\!\mathbf{s}'\!\!,\mathbf{a}')^T\!  \rbrace 
\end{align}
motivated by the fact that a continuous function over a compact set may be approximated uniformly by a function in a RKHS equipped with a universal kernel \cite{micchelli2006universal}. 

To apply the Representer Theorem, we require the cost to be coercive in $Q$ \cite{argyriou2009there}, which may be satisfied through use of a Hilbert-norm regularizer, so we define the regularized cost functional $J(Q) = L(Q) + (\lambda/2) \|Q\|_{\ccalH}^2$ and solve the regularized problem \eqref{eq:main_prob}, i.e.
\begin{align}  \label{eq:lossprob} 
   Q^{*} = \argmin_{Q\in \ccalH} J(Q) 
         = \argmin_{Q\in \ccalH} L(Q) + \frac{\lambda}{2} \|Q\|_{\ccalH}^2 .
\end{align}
Thus, finding a locally optimal action-value function in an MDP amounts to solving the RKHS-valued compositional stochastic program with a non-convex objective defined by the Bellman optimality equation \eqref{eq:bellop}. This action-value function can then be used to obtain the optimal policy \eqref{eq:pistar}. In the following section, we turn to iterative stochastic methods to solve \eqref{eq:lossprob}. 
We point out that this is a step back from the original intent of solving \eqref{eq:main_prob} to then find optimal policies $\pi^*$ using \eqref{eq:pistar}. This is the case because the assumption we have made about $Q^{*}$ being representable in the RKHS $\ccalH$ need not be true. More importantly, the functional $J(Q)$ is not convex in $Q$ and there is no guarantee that a local minimum of $J(Q)$ will be the optimal policy $Q^*$. This is a significant difference relative to policy evaluation problems \cite{pkgtd}.

\section{Stochastic Quasi-Gradient Method}\label{sec:alg}

To solve \eqref{eq:lossprob}, we propose applying a functional variant of stochastic quasi-gradient (SQG) descent to the loss function $J(Q)$ [cf. \eqref{eq:lossprob}]. The reasoning for this approach rather than a stochastic gradient method is the nested expectations cause the functional stochastic gradient to be still dependent on a second expectation which is not computable, and SQG circumvents this issue. Then, we apply the Representer Theorem \eqref{eq:qrkhs} (``kernel trick'') to obtain a parameterization of this optimization sequence, which has per-iteration complexity. We then mitigate this untenable complexity growth while preserving optimality using greedy compressive methods, inspired by \cite{polk,pkgtd}.

To find a stationary point of \eqref{eq:lossprob} we use quasi-gradients $\nabla_Q J(Q)$ of the functional $J(Q)$ relative to the function $Q$ in an iterative process. To do so, introduce an iteration index $t$ and let $Q_t$ be the estimate of the stationary point at iteration $t$. Further consider a random state action pair $(\bbs_t,\bba_t)$ independently and randomly chosen from the distribution $\mbP(d\bbs, d\bba)$. Action $\bba_t$ is executed from state $\bbs_t$ resulting in the system moving to state $\bbs'_t$. This outcome is recorded along with reward $r(\bbs_t, \bba_t, \bbs'_t)$ and the action $a'_t$ that maximizes the action-value function $Q_t$ when the system is in state $\bbs'_t$, i.e.,
\begin{equation}  \label{eq:actargmax2} 
   \bba'_t := \argmax_{a'} Q_t(\bbs_t',\bba').
\end{equation}
The state (S) $\bbs_t$, action (A) $\bba_t$, reward (R) $r(\bbs_t, \bba_t, \bbs'_t)$, state (S) $\bbs_t'$, action (A) $\bba_t'$ are collectively referred to as the SARSA tuple at time $t$.

Further consider the expressions for $J(Q)$ in \eqref{eq:lossprob} and $L(Q)$ in \eqref{eq:cost_functional} and exchange order of the expectation and differentiation operators to write the gradient of $J(Q)$ as 
\begin{align}  \label{eq:qfunctderiv2} 
   \nabla_Q J(Q_t) =  \mbE_{\bbs_t,\bba_t} \Big[ f(Q_t;\bbs_t,\bba_t) \times \nabla_Q f(Q_t;\bbs_t,\bba_t) \Big] 
                    + \lambda Q_t  .
\end{align}
To compute the directional derivative $\nabla_Q g(Q)$ in \eqref{eq:qfunctderiv2}, we need to address differentiation of the softmax and its approximation properties with respect to the exact maximum, which is done in the following remark.

\begin{remark}\label{remark_softmax}(Softmax Gradient Error)\normalfont
\ The functional derivative of \eqref{eq:softmax} takes the form
\begin{equation}\label{eq:softmax_derivative}
\nabla_Q \text{softmax}_{\bba' \in \ccalA} Q(\bbs', \bba') = \frac{\int_{\bba' \in \ccalA} e^{\eta Q(\bbs',\bba')} \kappa(\bbs', \bba', \cdot) d\bba }{\int_{\bba' \in \ccalA} e^{\eta Q(\bbs',\bba')} d\bba'}
\end{equation}
by applying Leibniz Rule, Chain Rule, and the reproducing property of the kernel. Moreover, a factor of $\eta$ cancels. Observe that as $\eta \rightarrow \infty$, the softmax becomes closer to the exact (hard) maximum, and the integrals in \eqref{eq:softmax_derivative} approach unit, and the only term that remains is  $\kappa(\bbs', \bba', \cdot)$. This term may be used in place of \eqref{eq:softmax_derivative} to avoid computing the integral, and yields the functional gradient of the exact maximum instead of the softmax. Doing so, however, comes at the cost of computing of the maximizer of the $Q$ function $\bba'$.
\end{remark}

Observe that to obtain samples of $ \nabla_Q J(Q,\mathbf{s},\mathbf{a},\mathbf{s}')$ we require two different queries to a simulation oracle: one to approximate the inner expectation over the Markov transition dynamics defined by $\bbs'$, and one for \emph{each initial pair} $\bbs, \bba$ which defines the outer expectation. This complication, called the ``double sampling problem," was first identified in \cite{ermoliev1983stochastic,borkar1997stochastic}, has been ameliorated through use of two time-scale stochastic approximation, which may be viewed as a stochastic variant of quasi-gradient methods \cite{wang2017stochastic}. 

Following this line of reasoning,  we build up the total expectation of one of the terms in \eqref{eq:qfunctderiv2} while doing stochastic descent with respect to the other. In principle, it is possible to build up the expectation of either term in \eqref{eq:qfunctderiv2}, but the mean of the difference of kernel evaluations is of infinite complexity. On the other hand, the \emph{temporal action difference}, defined as the difference between the action-value function evaluated at state-action pair $(\mathbf{s},\mathbf{a})$ and the action-value function evaluated at next state and the instantaneous maximizing action $(\mathbf{s}',\mathbf{a}')$:
\begin{equation}  \label{eq:tempactdiff} 
\delta := r(\mathbf{s},\mathbf{a},\mathbf{s}') + \gamma Q(\mathbf{s}',\mathbf{a}') - Q(\mathbf{s},\mathbf{a})
\end{equation}
is a \emph{scalar}, and thus so is its total expected value. Therefore, for obvious complexity motivations, we build up the total expectation of \eqref{eq:tempactdiff}. To do so, we propose recursively averaging realizations of \eqref{eq:tempactdiff} through the following auxiliary sequence $z_t$, initialized as null $z_0=0$:
\begin{align} \label{eq:zupdate}
  & \delta_t   := r(\mathbf{s}_t,\mathbf{a}_t,\mathbf{s}_t') + \gamma Q(\mathbf{s}_t',\mathbf{a}_t') - Q(\mathbf{s}_t,\mathbf{a}_t) \; , 
  \nonumber \\
  & z_{t+1} = (1 - \beta_t)z_t + \beta_t \delta_t 
\end{align}
where $(\mathbf{s}_t, \mathbf{a}_t, \mathbf{s}_t')$ is an independent realization of the random triple $(\mathbf{s},\mathbf{a},\mathbf{s}')$ and $\beta_t \in (0,1)$ is a learning rate. 

To define the stochastic descent step, we replace the first term inside the outer expectation in \eqref{eq:qfunctderiv2} with its instantaneous approximation $\lbrack\gamma \kappa((\mathbf{s}',\mathbf{a}'),\cdot) - \kappa((\mathbf{s},\mathbf{a}),\cdot) \rbrack$ evaluated at a sample triple $(\mathbf{s}_t, \mathbf{a}_t, \mathbf{s}_t')$, which yields the stochastic quasi-gradient step: 
\begin{align} \label{eq:qupdate}
\!\!\! Q_{t\!+\!1} \!\!\!=\!\! (1\!\! -\!\!\alpha_t\!\lambda\!)Q_t\!(\!\cdot\!)  
\!\!-\!\! \alpha_t(\!\gamma \kappa(\mathbf{s}_t'\!,\mathbf{a}_t'\!,\!\cdot\!) \!\!-\!\! \kappa(\mathbf{s}_t\!,\mathbf{a}_t\!,\!\cdot\!))z_{t\!+\!1}\!)
\end{align}
where the coefficient $(1-\alpha_t \lambda)$ comes from the regularizer and $\alpha_t$ is a positive scalar learning rate. Moreover, $\bba_t' = \argmax_\bbb Q_t(\bbs',\bbb)$ is the instantaneous $Q$-function maximizing action.  Now, using similar logic to \cite{kivinen2002online}, we may extract a tractable parameterization of the infinite dimensional function sequence \eqref{eq:qupdate}, exploiting properties of the RKHS \eqref{eq:repprop}.

\noindent\textbf{Kernel Parametrization} 
Suppose $Q_0 = 0 \in \ccalH$. Then the update in \eqref{eq:qupdate} at time $t$, inductively making use of the Representer Theorem, implies the function $Q_t$ is a kernel expansion of past state-action tuples $(\mathbf{s}_t, \mathbf{a}_t, \mathbf{s}_t')$
\begin{align} \label{eq:param}
{Q}_t(s,a) = \sum_{n=1}^{2(t-1)} w_n \kappa(\mathbf{v}_n,(\mathbf{s},\mathbf{a})) = \mathbf{w}_t^T \mathbf{\kappa}_{\mathbf{X_t}}((\mathbf{s},\mathbf{a}))
\end{align}
The kernel expansion in \eqref{eq:param}, together with the functional update \eqref{eq:qupdate}, yields the fact that functional SQG in $\ccalH$ amounts to updating the kernel dictionary $\mathbf{X}_{t}\in \mathbb{R}^{p \times 2(t-1)}$ and coefficient vector $\mathbf{w}_t \in \mathbb{R}^{2(t-1)} $ as
\begin{align} \label{eq:vectorupdate}
\mathbf{X}_{t+1} = \lbrack \mathbf{X}_t, (\mathbf{s}_t,\mathbf{a}_t), (\mathbf{s}_t',\mathbf{a}_t')  \rbrack \;, \;\;\;\;\; 
\mathbf{w}_{t+1} = \lbrack (1-\alpha_t \lambda) \mathbf{w}_t, \alpha_t z_{t+1}, -\alpha_t \gamma z_{t+1}\rbrack
\end{align}
In \eqref{eq:vectorupdate}, the coefficient vector $\mathbf{w}_t \in \mathbb{R}^{2(t-1)}$ and dictionary $\mathbf{X}_t \in \mathbb{R}^{p \times 2(t-1)}$ are defined as
\begin{align} \label{eq:wxparam}
\mathbf{w}_t =  \lbrack w_1, \ldots, w_{2(t-1)}\rbrack 
\;, \;\;\;\;\; 
\mathbf{X}_t =  \lbrack (\mathbf{s}_1,\mathbf{a}_1), (\mathbf{s}_1',\mathbf{a}_1'), \ldots,  (\mathbf{s}_{t-1},\mathbf{a}_{t-1}),(\mathbf{s}_{t-1}',\mathbf{a}_{t-1}')\rbrack,  
\end{align}
and in \eqref{eq:param}, we introduce the notation $\mathbf{v}_n = (\mathbf{s}_n, \mathbf{a}_n)$ for $n$ even and $\mathbf{v}_n = (\mathbf{s}_n', \mathbf{a}_n')$ for $n$ odd. Moreover, in \eqref{eq:param}, we make use of a concept called the empirical kernel map associated with dictionary $\mathbf{X}_t$, defined as
\begin{align} 
\mathbf{\kappa}_{\mathbf{X}_t} (\cdot)& = \lbrack (\kappa((\mathbf{s}_1,\mathbf{a}_1),\cdot), \kappa((\mathbf{s}_1',\mathbf{a}_1'),\cdot),  \ldots,\kappa((\mathbf{s}_{t-1},\mathbf{a}_{t-1}),\cdot),\kappa((\mathbf{s}_{t-1}',\mathbf{a}_{t-1}'),\cdot)\rbrack^T \; .
\end{align}
Observe that \eqref{eq:vectorupdate} causes 
$\mathbf{X}_{t+1}$ to have two more columns than its predecessor $\mathbf{X}_t$. We define the \textit{model order} as the number of data points (columns) $M_t$ in the dictionary at time t, which for functional stochastic quasi-gradient descent is $M_t = 2(t-1)$. Asymptotically, then, the complexity of storing ${Q}_t(\cdot)$ is infinite, and even for moderately large training sets is untenable. Next, we address this intractable complexity blowup, inspired by \cite{polk,pkgtd}, using greedy compression methods \cite{mallat1993matching}.

\noindent\textbf{Sparse Stochastic Subspace Projections} 
Since the update step \eqref{eq:qupdate} has complexity $\mathcal{O}(t)$ due to the RKHS parametrization, it is impractical in settings with streaming data or arbitrarily large training sets. We address this issue by replacing the stochastic quasi-descent step \eqref{eq:qupdate} with an orthogonally projected variant, where the projection is onto a low-dimensional functional subspace of the RKHS $\ccalH_{\textbf{D}_{t+1}} \subset \ccalH$ 
\begin{align}\label{eq:proj_fsqg}
Q_{t+1} = \mathcal{P}_{\ccalH_{\mathbf{D}_{t+1}}} \lbrack (1-\alpha_t\lambda){Q}_t(\cdot) 
 - \alpha_t(\gamma \kappa(\mathbf{s}_t',\mathbf{a}_t',\cdot) \! - \!\kappa(\mathbf{s}_t,\mathbf{a}_t,\cdot))z_{t+1}) \rbrack 
\end{align}
where $\ccalH_{\mathbf{D}_{t+1}} = \text{span}  \{((\mathbf{s}_n,\mathbf{a}_n),\cdot) \}_{n=1}^{M_t}$ for some collection of sample instances $\lbrace (\mathbf{s}_n,\mathbf{a}_n) \rbrace \subset \lbrace (\mathbf{s}_t, \mathbf{a}_t) \rbrace_{u \leq t}$. We define $\mathbf{\kappa}_\mathbf{D}(\cdot) = \lbrace \kappa((\mathbf{s}_1,\mathbf{a}_1),\cdot) \ldots \kappa((\mathbf{s}_M,\mathbf{a}_M),\cdot)  \rbrace$ and $\kappa_{\mathbf{D},\mathbf{D}}$ as the resulting kernel matrix from this dictionary. We seek function parsimony by selecting dictionaries $\mathbf{D}$ such that $M_t << \mathcal{O}(t)$. Suppose that $Q_t$ is parameterized by model points $\bbD_t$ and weights $\bbw_t$. Then, we denote $\tilde{Q}_{t+1}(\cdot) = (1-\alpha_t\lambda){Q}_t(\cdot) - \alpha_t(\gamma \kappa(\mathbf{s}_t',\mathbf{a}_t',\cdot) - \kappa(\mathbf{s}_t,\mathbf{a}_t,\cdot))z_{t+1})$ as the SQG step without projection. This may be represented by dictionary and weight vector [cf. \eqref{eq:vectorupdate}]:  
\begin{align} \label{eq:dvectorupdate}
\tilde{\mathbf{D}}_{t+1} = \lbrack \mathbf{D}_t, (\mathbf{s}_t,\mathbf{a}_t), (\mathbf{s}_t',\mathbf{a}_t')  \rbrack \;,  
\;\;\;\;\;
\tilde{\mathbf{w}}_{t+1} = \lbrack (1-\alpha_t \lambda) \mathbf{w}_t, \alpha_t z_{t+1}, -\alpha_t \gamma z_{t+1}\rbrack \; ,
\end{align}
where $z_{t+1}$ in \eqref{eq:dvectorupdate} is computed by \eqref{eq:zupdate} using $Q_t$ obtained from \eqref{eq:proj_fsqg}:
\begin{align} \label{eq:zupdateproj}
   \delta_t   := r(\mathbf{s}_t,\mathbf{a}_t,\mathbf{s}_t') + \gamma Q_t(\mathbf{s}_t',\mathbf{a}_t') - Q_t(\mathbf{s}_t,\mathbf{a}_t) \; , 
\;\;\;\;\;
   z_{t+1} = (1 - \beta_t)z_t + \beta_t \delta_t \; .
\end{align}
Observe that $\tilde{\mathbf{D}}_{t+1}$ has $\tilde{M}_{t+1} = M_{t} +2$ columns which is the length of $\tilde{\mathbf{w}}_{t+1}$.
We proceed to describe the construction of the subspaces $\ccalH_{\bbD_{t+1}}$ onto which the SQG iterates are projected in \eqref{eq:proj_fsqg}. 
Specifically, we select the kernel dictionary $\mathbf{D}_{t+1}$ via greedy compression. We form $\mathbf{D}_{t+1}$ by selecting a subset of $M_{t+1}$ columns from $\tilde{\mathbf{D}}_{t+1}$ that best approximates $\tilde{Q}_{t+1}$ in terms of Hilbert norm error. To accomplish this, we use kernel orthogonal matching pursuit \cite{polk,pkgtd} with error tolerance $\epsilon_t$ to find a compressed dictionary $\mathbf{D}_{t+1}$ from  $\tilde{\mathbf{D}}_{t+1}$, the one that adds the latest samples.
For a fixed dictionary $\mathbf{D}_{t+1}$, the update for the kernel weights is a least-squares problem on the coefficient vector:
\begin{equation} \label{eq:wupdate}
\mathbf{w}_{t+1} = \kappa_{\mathbf{D}_{t+1}\mathbf{D}_{t+1}} ^{-1}  \kappa_{\mathbf{D}_{t+1}\tilde{\mathbf{D}}_{t+1}} \tilde{\mathbf{w}}_{t+1}
\end{equation}
We tune $\epsilon_t$ to ensure both stochastic descent and finite model order -- see the next section. 

We summarize the proposed method, KQ-Learning, in Algorithm \ref{qlearning}, the execution of the stochastic projection of the functional SQG iterates onto subspaces $\ccalH_{\mathbf{D}_{t+1}}$.
We begin with a null function $Q_0 = 0$, i.e., empty dictionary and coefficients (Step 1). At each step, given an i.i.d. sample $(\mathbf{s}_t, \mathbf{a}_t, \mathbf{s}_t')$ and step-size $\alpha_t$, $\beta_t$ (Steps 2-5), we compute the unconstrained functional SQG iterate  $\tilde{Q}_{t+1}(\cdot) = (1-\alpha_t\lambda){Q}_t(\cdot) - \alpha_t(\gamma \kappa(\mathbf{s}_t',\mathbf{a}_t',\cdot) - \kappa(\mathbf{s}_t,\mathbf{a}_t,\cdot))z_{t+1})$ parametrized by $\tilde{\mathbf{D}}_{t+1}$ and $\tilde{\mathbf{w}}_{t+1}$ (Steps 6-7), which are fed into KOMP (Algorithm 2) \cite{polk} with budget $\epsilon_t$, (Step 8). KOMP then returns a lower complexity estimate  $Q_t$ of $\tilde{Q_t}$ that is $\eps_t$ away in $\ccalH$.
 \begin{algorithm}[t]
 \caption{KQ-Learning}\label{qlearning}
\begin{algorithmic}[1]
 \renewcommand{\algorithmicrequire}{\textbf{Input:}}
 \renewcommand{\algorithmicensure}{\textbf{Output:}}
 \REQUIRE $C, \{ \alpha_t, \beta_t \}_{t=0,1,2\ldots}$
  \STATE $Q_0(\cdot) = 0, D_0=[], w_0=[], z_0 = 0$
  \FOR {$t = 0,1,2,\ldots$}
  \STATE Obtain sample $(\mathbf{s}_t,\mathbf{a}_t,\mathbf{s}_t')$ via exploratory policy
\STATE Compute maximizing action \\ $\mathbf{a}' = \argmax_{\bba}  Q_t(\mathbf{s}_t',\bba)$
\STATE Update temporal action diff. $\delta_t$ and aux. seq. $z_{t+1}$ \\
 $\delta_t = r(\mathbf{s}_t,\mathbf{a}_t,\mathbf{s}_t') + \gamma  Q_t(\mathbf{s}_t',\mathbf{a}_t') - Q_t(\mathbf{s}_t,\mathbf{a}_t)$  \\
 $z_{t+1} = (1-\beta_t)z_t + \beta_t\delta_t \; .$
\STATE Compute functional stochastic quasi-grad. step \\
$\!\tilde{Q}_{t+1}\! =  \!(\!1\!-\!\alpha_t\lambda\!\!){Q}_t \!-\! \alpha_t z_{t+1}(\!\gamma \kappa(\!\mathbf{s}_t'\!,\mathbf{a}_t'\!,\!\cdot\!) \!-\! \kappa(\mathbf{s}_t,\mathbf{a}_t,\!\cdot\!)\!).$
\STATE Update dictionary $\tilde{D}_{t+1} = \lbrack D_t,(\mathbf{s},\mathbf{a}),(\mathbf{s}',\mathbf{a}') \rbrack $ ,\\ weights $\tilde{w}_{t+1} = \lbrack(1 - \alpha_t \lambda) w_t , \alpha_t z_{t+1} , -\alpha_t \gamma z_{t+1}\rbrack$.
\STATE Compress function using KOMP with budget $\epsilon_t = C \alpha_t^2$
\\ $ (Q_{t+1},D_{t+1}, w_{t+1}) = \mathbf{KOMP}(\tilde{Q}_{t+1},\tilde{D}_{t+1}, \tilde{w}_{t+1}, \epsilon_t) $
  \ENDFOR
 \RETURN $Q$ 
 \end{algorithmic} 

\end{algorithm}
  \begin{algorithm}[t]
 \caption{Destructive Kernel Orthogonal Matching Pursuit (KOMP)}\label{komp}
\begin{algorithmic}[1]
 \renewcommand{\algorithmicrequire}{\textbf{Input:}}
 \renewcommand{\algorithmicensure}{\textbf{Output:}}
 \REQUIRE function $\tilde{Q}$ defined by dict $\tilde{D} \in \mathbb{R}^{p \times \tilde{M}}$,  $\tilde{w} \in \mathbb{R}^{\tilde{M}}$, approx. budget $\epsilon_t > 0$
 \\ \textit{Initialize} : $Q = \tilde{Q}$, dictionary $D = \tilde{D}$ with indices $\mathcal{I}$, model order $M = \tilde{M}$, coeffs $w = \tilde{w}$.
  \WHILE {candidate dictionary is non-empty $\mathcal{I} \neq \emptyset $}
  \FOR {$j = 1, \ldots, \tilde{M}$}
  \STATE Find minimal approximation error with dictionary element $d_j$ removed \\  
  $\hspace{-5mm} \gamma_j = \min_{w_{\mathcal{I} \backslash \lbrace j \rbrace } \in \mathbb{R}^{M-1}} \| \tilde{Q}(\cdot) - \sum_{k \in \mathcal{I}\backslash \lbrace j \rbrace } w_k \kappa(d_k, \cdot) \|_{\ccalH}
$
\ENDFOR
  \STATE Find dictionary index minimizing approximation error : $j^* = \argmin_{j \in \mathcal{I}} \gamma_j$
  \IF {minimal approximation error exceeds threshold $\gamma_{j^*} > \epsilon_t$}
  	\BREAK
  \ELSE
  \STATE {Prune dictionary $D \leftarrow D_{\mathcal{I} \backslash \lbrace j^* \rbrace}$}
  \STATE {Revise set $\mathcal{I} \!\!\! \leftarrow \!\! \mathcal{I} \backslash \lbrace j^* \rbrace$} and model order $M \!\! \leftarrow \!\! M\!\!-\!\!1$
  \STATE Compute updated weights $w$ defined by the current dictionary $D$ \\
  $ w = \argmin_{w \in \mathbb{R}^M } \|\tilde{Q}(\cdot) - w^T \kappa_D(\cdot) \|_{\ccalH} $
  \ENDIF 
  \ENDWHILE
 \RETURN $V,D,w$ of model order $M \leq \tilde{M}$ such that $\|Q - \tilde{Q} \|_\ccalH \leq \epsilon_t$
 \end{algorithmic} 
 \end{algorithm}

 \section{Convergence Analysis}\label{sec:convergence}
 
In this section, we shift focus to the task of establishing that the sequence of action-value function estimates generated by Algorithm \ref{qlearning} actually yield a locally optimal solution to the Bellman optimality equation, which, given intrinsic the non-convexity of the problem setting, is the best one may hope for in general through use of numerical stochastic optimization methods. Our analysis extends the ideas of coupled supermartingales in reproducing kernel Hilbert spaces \cite{pkgtd}, which have been used to establish convergent policy evaluation approaches in infinite MDPs (a convex problem), to non-convex settings, and further generalizes the non-convex vector-valued setting of \cite{wang2017stochastic}. 

Before proceeding with the details of the technical setting, we introduce a few definitions which simplify derivations greatly. In particular, for further reference, we use \eqref{eq:actargmax2} to define   $\mathbf{a}_t' = \argmax_{\bba} Q_t(\mathbf{s}_t',\bba)$, the instantaneous maximizer of the action-value function and defines the direction of the gradient.  We also define the functional stochastic quasi-gradient of the regularized objective
\begin{align} \label{eq:unprojectedgradproof}
\hat{\nabla}_Q  J(Q_t, z_{t+1}; \mathbf{s}_t,\mathbf{a}_t,\mathbf{s}_t') =   (\gamma \kappa((\mathbf{s}_t',\mathbf{a}_t'),\cdot) 
-\kappa((\mathbf{s}_t,\mathbf{a}_t),\cdot))z_{t+1} + \lambda Q_t
\end{align}
 and its sparse-subspace projected variant as
 \begin{align} \label{eq:projectedgradproof}
  \tilde{\nabla}_Q J(Q_t, z_{t+1}; \mathbf{s}_t,\mathbf{a}_t,\mathbf{s}_t') =    (Q_t \!\! - \!\! \mathcal{P}_{\ccalH_{D_{t+1}}} \!\! \lbrack Q_t - {\alpha}_t \!\! \hat{\nabla \!}_Q J(Q_t, z_{t+1}; \mathbf{s}_t,\mathbf{a}_t,\mathbf{s}_t') \rbrack ) \! / \! \alpha_t
 \end{align}
 Note that the update may be rewritten as a stochastic projected quasi-gradient step rather than a stochastic quasi-gradient step followed by a set projection, i.e.,
 \begin{equation} \label{eq:qprojectedupdateproof}
 Q_{t+1} = Q_t - {\alpha}_t  \tilde{\nabla}_Q J(Q_t, z_{t+1}; \mathbf{s}_t,\mathbf{a}_t,\mathbf{s}_t')
 \end{equation}
With these definitions, we may state our main assumptions required to establish convergence of Algorithm \ref{qlearning}.
\begin{assumption}\label{as:compact}
The state space  $\mathcal{S} \subset \mathbb{R}^p$ and action space $\mathcal{A} \subset \mathbb{R}^q$ are compact, and the reproducing kernel map may be bounded as
\begin{equation} \label{eq:boundedkernelproof}
\sup_{\mathbf{s} \in \mathcal{S}, \mathbf{a} \in \mathcal{A}} \sqrt{\kappa((\mathbf{s},\mathbf{a}),(\mathbf{s},\mathbf{a}))} = K < \infty
\end{equation}
Moreover, the subspaces $\ccalH_{\bbD_t}$ are intersected with some finite Hilbert norm ball for each $t$.
\end{assumption}
\begin{assumption}\label{as:mean_variance_delta}
 The temporal action difference $\delta$ and auxiliary sequence $z$ satisfy the zero-mean, finite conditional variance, and Lipschitz continuity conditions, respectively,
\begin{equation} \label{eq:mean_variance_delta}
\!\!\! \mathbb{E} \lbrack \delta \! \mid \! \mathbf{s},\mathbf{a} \rbrack = \bar{\delta} \; , \ 
 \mathbb{E} \lbrack (\delta \! - \! \bar{\delta})^2\rbrack \! \leq \! {\sigma}_{\delta}^2 \; ,\
 \mathbb{E} \lbrack z^2 \! \mid \! \mathbf{s}, \mathbf{a} \rbrack \! \leq \! G_{\delta}^2  
\end{equation}
where $\sigma_{\delta}$ and $G_{\delta}$ are positive scalars, and $\bar{\delta} = \mathbb{E} \lbrace \delta \mid \mathbf{s}, \mathbf{a} \rbrace$ is the expected value of the temporal action difference conditioned on the state $\bbs$ and action $\mathbf{a}$.
\end{assumption}
\begin{assumption}\label{as:mean_variance_q}The functional gradient of the temporal action difference is an unbiased estimate for $\nabla_Q J(Q)$ and the difference of the reproducing kernels expression has finite conditional variance:
\begin{equation} \label{eq:finitevariance1proof}
\mathbb{E} \lbrack (\gamma \kappa((\mathbf{s}_t',\mathbf{a}_t'),\cdot) - \kappa((\mathbf{s}_t,\mathbf{a}_t),\cdot))\delta \rbrack = \nabla_Q J(Q)
\end{equation}
\begin{equation} \label{eq:finitevariance2proof}
\mathbb{E} \lbrace \| \gamma \kappa((\mathbf{s}_t',\mathbf{a}_t'),\cdot) - \kappa((\mathbf{s}_t,\mathbf{a}_t),\cdot) \|^2_{\ccalH}|\mathcal{F}_t\rbrace \leq G_Q^2
\end{equation}
Moreover, the projected stochastic quasi-gradient of the objective has finite second conditional moment as 
\begin{equation} \label{eq:finitemomentproof}
\mathbb{E} \lbrace \|  \tilde{\nabla}_Q J(Q_t, z_{t+1}; \mathbf{s}_t,\mathbf{a}_t,\mathbf{s}_t') \|^2_{\ccalH}|\mathcal{F}_t\rbrace \leq {\sigma}_Q^2
\end{equation}
and the temporal action difference is Lipschitz continuous with respect to the action-value function Q. Moreover, for any two distinct $\delta$ and $\bar{\delta}$, we have 
\begin{equation} \label{eq:deltaqproof}
\|\delta - \bar{\delta}\| \leq L_Q \| Q - \tilde{Q} \|_{\ccalH}
\end{equation}
with $Q$, $\tilde{Q} \in \ccalH$ distinct $Q$-functions; $L_Q > 0$ is a scalar.
\end{assumption}

Assumption \ref{as:compact} regarding the compactness of the state and action spaces of the MDP holds for most application settings and limits the radius of the set from which the MDP trajectory is sampled. The mean and variance properties of the temporal difference stated in Assumption \ref{as:mean_variance_delta} are necessary to bound the error in the descent direction associated with the stochastic sub-sampling and are required to establish convergence of stochastic methods. Assumption \ref{as:mean_variance_q} is similar to Assumption \ref{as:mean_variance_delta}, but instead of establishing bounds on the stochastic approximation error of the temporal difference, limits stochastic error variance in the RKHS. The term related to the maximum of the $Q$ function in the temporal action difference is Lipschitz in the infinity norm since $Q$ is automatically Lipschitz since it belongs to the RKHS. Thus, this term can be related to the Hilbert norm through a constant factor. Hence, \eqref{eq:deltaqproof} is only limits how non-smooth the reward function may be. These are natural extensions of the conditions needed for vector-valued stochastic compositional gradient methods.

Due to Assumption \ref{as:compact} and the use of set projections in \eqref{eq:proj_fsqg}, we have that $Q_t$ is always bounded in Hilbert norm, i.e., there exists some $0<D<\infty$ such that
 \begin{equation} \label{eq:compactnessproof}
 \|Q_t\|_{\ccalH} \leq D \text{ for all } t \; .
 \end{equation}

With these technical conditions, we can derive a coupled stochastic descent-type relationship of Algorithm \ref{qlearning} and then apply the Coupled Supermartingale Theorem \cite{wang2013incremental}[Lemma 6] to establish convergence, which we state next.
\begin{theorem}\label{main_theorem}
Consider the sequence $z_t$ and $\lbrace Q_t\rbrace$ as stated in Algorithm \ref{qlearning}. Assume the regularizer is positive $\lambda > 0$, Assumptions \ref{as:compact}-\ref{as:mean_variance_q} hold, and the step-size conditions hold, with $C>0$ a positive constant:\vspace{-1.5mm}
\begin{align} \label{eq:theorem1proof}
\sum_{t=1}^\infty \alpha_t  = \infty, \sum_{t=1}^\infty \beta_t  = \infty, \sum_{t=1}^\infty \alpha_t^2 + \beta_t^2 + \frac{\alpha_t^2}{\beta_t}  < \infty , \epsilon_t = C \alpha_t^2 
\end{align}
Then $ \|\nabla_Q J(Q) \|_{\ccalH}$ converges to null with probability 1, and hence $Q_t$ attains a stationary point of \eqref{eq:lossprob}. In particular, the limit of $Q_t$ achieves the regularized  Bellman fixed point restricted to the RKHS.  
\end{theorem}
%

See Appendix B.

Theorem \ref{main_theorem} establishes that Algorithm \ref{qlearning} converges almost surely to a stationary solution of the problem \eqref{eq:lossprob} defined by the Bellman optimality equation in a continuous MDP. This is one of the first Lyapunov stability results for $Q$-learning in continuous state-action spaces with nonlinear function parameterizations, which are intrinsically necessary when the $Q$-function does not admit a lookup table (matrix) representation, and should form the foundation for value-function based reinforcement learning in continuous spaces. A key feature of this result is that the complexity of the function parameterization will not grow untenably large due to the use of our KOMP-based compression method which ties the sparsification bias $\eps_t$ to the algorithm step-size $\alpha_t$. In particular, by modifying the above exact convergence result for diminishing learning rates to one in which they are kept constant, we are able to keep constant compression budgets as well, and establish convergence to a neighborhood as well as the finiteness of the model order of  $Q$, as we state next.
\begin{theorem}\label{main_theorem2}
Consider the sequence $z_t$ and $\lbrace Q_t\rbrace$ as stated in Algorithm \ref{qlearning}. Assume the regularizer is positive $\lambda > 0$, Assumptions \ref{as:compact}-\ref{as:mean_variance_q} hold, and the step-sizes are chosen as constant such that $0 < \alpha < \beta < 1$, with $\eps=C \alpha^2$ and the parsimony constant $C>0$ is positive. Then the Bellman error converges to a neighborhood in expectation, i.e.:
\begin{align} \label{eq:theorem2}
\liminf_{t\rightarrow \infty} \ \mathbb{E}[J(Q_t)] &\leq \ccalO\!\!\left(\! \!\!\frac{\alpha \beta}{\beta - \alpha} \!\!\! \left[\!\!1 \!+\! \sqrt{\!\!1\! +\! \frac{\beta - \alpha}{\alpha \beta}\!\!\!\left(\!\frac{1}{\beta} \!+ \!\frac{\beta^2}{\alpha^2}  \!\!\right)} \!\!\! \right]\!\!   \right)
\end{align}
\end{theorem}
%
See Appendix C. 

The expression on the right-hand side of \eqref{eq:theorem2} is a complicated posynomial of $\alpha$ and $\beta$, but is positive provided $\beta>\alpha$, and for a fixed $\beta$ increases as $\alpha$ increases. This means that more aggressive selections of $\alpha$, for a given $\beta$, yield a larger limiting lower bound on the Bellman error. A simple example which satisfies the constant step-size conditions $0 < \alpha < \beta < 1$ is $\beta = \alpha + \iota$ for some small constant $\iota >0$. This is consistent with the diminishing step-size conditions where $\alpha_t/\beta_t \rightarrow 0$ means that $\alpha_t$ must be smaller than $\beta_t$ which is in $(0,1)$. 

An additional salient feature of the parameter choice given in Theorem \ref{main_theorem2} is that \cite{pkgtd}[Corollary 1] applies, and thus we may conclude that the $Q$-function parameterization is at-worst finite during learning when used with constant step-sizes and compression budget.
In subsequent sections, we investigate the empirical validity of the proposed approach on two autonomous control tasks: the Inverted Pendulum and Continuous Mountain Car, for which observe consistent convergence in practice. To do so, first some implementation details of Algorithm \ref{qlearning} must be addressed.

\begin{remark}\normalfont
Observe that \eqref{eq:tdupdate} bears a phantom resemblance to Watkins' Q-Learning algorithm \cite{Watkins1989}; however, it is unclear how to extend \cite{Watkins1989} to continuous MDPs where function approximation is required. In practice, using \eqref{eq:tdupdate} for all updates, we observe globally steady policy learning and convergence of Bellman error, suggesting a link between \eqref{eq:tdupdate} and stochastic fixed point methods \cite{tsitsiklis1994asynchronous,jaakkola1994convergence}. This link is left to future investigation. For now, we simply note that stochastic fixed point iteration is fundamentally different than stochastic descent methods which rely on the construction of supermartingales, so results from the previous section do not apply to \eqref{eq:tdupdate}. Moreover, this update has also been referred to as  a temporal difference ``semi-gradient" update in Chapters 9-10 of \cite{sutton2018reinforcement}. 
\end{remark}

\section{Practical Considerations}\label{sec:approx}

The convergence guarantees for Algorithm \ref{qlearning} require sequentially observing state-action-next-state triples $(\bbs_t, \bba_t, r_t,  \bbs_t')$ independently and identically distributed. Doing so, however, only yields convergence toward a stationary point, which may or may not be the optimal $Q$ function. To improve the quality of stationary points to which we converge, it is vital to observe states that yield reward (an instantiation of the explore-exploit tradeoff). To do so, we adopt a standard practice in reinforcement learning which is to bias actions towards states that may accumulate more reward. 

The method in which we propose to bias actions is by selecting them according to the current estimate of the optimal policy, i.e., the greedy policy. However, when doing so, the KQ-Learning updates \eqref{eq:qupdate} computed using greedy samples $(\bbs_t, \bba_t, r_t,  \bbs_t')$ are composed of two points nearby in $\ccalS \times \ccalA$ space. These points are then evaluated by kernels and given approximately equal in opposite weight. Thus, this update is immediately pruned away during the execution of KOMP \cite{mallat1993matching,Vincent2002,polk}. In order to make use of the greedy samples and speed up convergence, we project the functional update onto just one kernel dictionary element, resulting in the update step:
\begin{equation} \label{eq:tdupdate}
\tilde{Q}_{t+1}\! =  \!(\!1\!-\!\alpha_t\lambda\!\!){Q}_t(\!\cdot\!) +\! \alpha_t(\kappa(\mathbf{s}_t,\mathbf{a}_t,\!\cdot\!)\!)z_{t+1}\!)
\end{equation}
The resulting procedure is summarized as Algorithm \ref{greedy}. First, trajectory samples are obtained using a greedy policy. Then, the temporal-action difference is computed and averaged recursively. Finally, we update the Q function via \eqref{eq:tdupdate} and compress it using Algorithm \ref{komp}. 

{\bf $\rho$-Greedy Actions and Hybrid Update}  To address the explore-exploit trade-off, we use an $\rho$-greedy policy \cite{singh2000convergence}: with probability $\rho$ we select a random action, and select a greedy action with probability $1-\rho$. We adopt this approach with $\rho$ decreasing linearly during training, meaning that as time passes more greedy actions are taken.

The algorithm when run with a  $\rho$-greedy policy is described as the \emph{Hybrid} algorithm, which uses Algorithm \ref{qlearning} when exploratory actions are taken and Algorithm \ref{greedy} for greedy actions. Practically, we find it useful to judiciously use training examples, which may be done with a data buffer. Thus, the hybrid algorithm is as follows: First, we accumulate trajectory samples in a buffer. Along with the $(\bbs_t, \bba_t, r_t, \bbs'_t)$ sample, we store an indicator whether $\bba_t$ was an exploratory action or greedy with respect to $Q_{t-1}$.  Then, samples are drawn at random from the buffer for training. We explore two different methods for obtaining samples from the buffer: uniformly at random, and prioritized sampling, which weighs each sample in the buffer by its observed Bellman error. For greedy actions, we use the update in \eqref{eq:tdupdate}, and for exploratory actions, we use the KQ-learning update from \ref{qlearning}. Finally, we use KOMP to compress the representation of the $Q$ function. 

{\bf Maximizing the $Q$ Function} In order to implement Algorithm \ref{greedy}, we apply simulated annealing \cite{simulatedannealing} to evaluate the instantaneous maximizing action $\bba_t = \argmax_\bba Q_t((\bbs_t, \bba)$. For a general reproducing kernel $\kappa(\cdot,\cdot)$, maximizing over a weighted sum of kernels is a non-convex optimization problem, so we get stuck in undesirable stationary points \cite{carreira2003number}. To reduce the chance that this undesirable outcome transpires, we use simulated annealing. First, we sample actions $\bba$ uniformly at random from the action space. Next, we use gradient ascent to refine our estimate of the global maximum of $Q$ for state $\bbs$. We use the Gaussian Radial Basis Function(RBF) kernel \eqref{eq:rbfdef}, so the $Q$ function is differentiable with respect to an arbitrary action $\mathbf{a}$:\vspace{-2mm}
\begin{equation}
(\nabla_{\mathbf{a}} Q)(\mathbf{s},\mathbf{a}) = Q(\mathbf{s},\mathbf{a}) \sum_{m=1}^M  w_m \Sigma_\mathbf{a} (\mathbf{a}-\mathbf{a}_m)^T
\end{equation}
and that gradient evaluations are cheap: typically their complexity scales with the model order of the $Q$ function which is a kept under control using Algorithm \ref{komp}.

 \begin{algorithm}[t]
 \caption{Semi-Gradient Greedy KQ-Learning}\label{greedy}
\begin{algorithmic}[1]
 \renewcommand{\algorithmicrequire}{\textbf{Input:}}
 \renewcommand{\algorithmicensure}{\textbf{Output:}}
 \REQUIRE $C, \{ \alpha_t, \beta_t\}_{t=0,1,2\ldots}$
  \STATE $Q_0(\cdot) = 0, D_0=[], w_0=[], z_0 = 0$
  \FOR {$t = 0,1,2,\ldots$}
  \STATE Obtain sample $(\mathbf{s}_t,\mathbf{a}_t,\mathbf{s}_t')$ via greedy policy 
\STATE Compute maximizing action: \\ $\bba'_t \!\!\!=\!\! \pi_t(\!\bbs_t')\! =\! \argmax_{\bba}  \!\!Q_t(\!\mathbf{s}_t',\bba)$
\STATE Update temporal action diff. $\delta_t$ and aux. seq. $z_{t+1}$ \\
 $\delta_t = r(\mathbf{s}_t,\mathbf{a}_t,\mathbf{s}_t') + \gamma  Q_t(\mathbf{s}_t',\mathbf{a}_t') - Q_t(\mathbf{s}_t,\mathbf{a}_t)$ \\
 $z_{t+1} = (1-\beta_t)z_t + \beta_t\delta_t \; .$

\STATE Compute update step \\
$\!\tilde{Q}_{t+1}\! =  \!(\!1\!-\!\alpha_t\lambda\!\!){Q}_t(\!\cdot\!) +\! \alpha_t z_{t+1} \kappa(\mathbf{s}_t,\mathbf{a}_t,\!\cdot\!).$
\STATE Update dictionary $\tilde{D}_{t+1} = \lbrack D_t,(\mathbf{s},\mathbf{a}) \rbrack $ ,\\ weights $\tilde{w}_{t+1} = \lbrack(1 - \alpha_t \lambda) w_t , \alpha_t z_{t+1} \rbrack$.

\STATE Compress function using KOMP with budget $\epsilon_t = C \alpha_t^2$
\\ $ (Q_{t+1},D_{t+1}, w_{t+1}) = \mathbf{KOMP}(\tilde{Q}_{t+1},\tilde{D}_{t+1}, \tilde{w}_{t+1}, \epsilon_t) $
  \ENDFOR
 \RETURN $Q$ 
 \end{algorithmic} 

\end{algorithm}
\begin{table*}[h]
\centering 
\renewcommand{\arraystretch}{1.07}
\begin{tabular}{l l l l l l l l l l l l l l}
\Xhline{2\arrayrulewidth}
& Environment & Algorithm & Replay Buffer & Policy & Steps & $\alpha$ & $\beta$ & $C$ & Kernel $\Sigma$            &   Order &  Loss &  Rewards   \\ 
\hline 
1 & Inv. Pendulum    & KQ        & Yes           & Exploratory       & 100K  & 0.25     & 1.00    & 2.00       & {[}0.5,0.5,2,0.5{]} & 137.17 & 0.79 & -1194.35                 \\
2 & Inv. Pendulum    & KQ        & Yes           & $\rho$-greedy & 100K  & 0.25     & 1.00    & 2.00       & {[}0.5,0.5,2,0.5{]} & 111.71 & 22.26 & -1493.65                \\
3 & Inv. Pendulum    & Hybrid    & Yes           & $\rho$-greedy & 500K  & 0.25     &  1.00    & 2.00       & {[}0.5,0.5,2,0.5{]} & {\bf 636.2 } & {\bf 0.99 } & {\bf -160.01  }                \\
4 & Inv. Pendulum    & SG        & Yes           & $\rho$-greedy & 200K  & 0.25     & 1.00    & 2.00       & {[}0.5,0.5,2,0.5{]} & {\bf 749.75 } & {\bf 2.92} & {\bf -150.36 }                \\
5 & Inv. Pendulum    & KQ        & No            & Exploratory       & 100K  & 0.25     & 1.00    & 2.00       & {[}0.5,0.5,2,0.5{]} & 134.14 & 0.72 & -1258.43                 \\
6 & Inv. Pendulum    & KQ        & No            & $\rho$-greedy & 100K  & 0.25     & 1.00    & 2.00       & {[}0.5,0.5,2,0.5{]} & 257.71 & 14.5 & -1258.45                 \\
7 & Inv. Pendulum    & Hybrid    & No            & $\rho$-greedy & 500K  & 0.25     & 1.00    & 2.00       & {[}0.5,0.5,2,0.5{]} & 684.39 & 0.61 & -180.37                  \\
8 & Inv. Pendulum    & SG        & No            & $\rho$-greedy & 200K  & 0.25     & 1.00    & 2.00       & {[}0.5,0.5,2,0.5{]} & 772.69 & 1.88 & -247.17                 \\
9 & Cont. M. Car  & KQ        & Prioritized   & Exploratory       & 100K  & 0.25     & 1.00    & 0.10       & {[}0.8,0.07,1.0{]}  & 44.54 & 0.41 & -20.61                    \\
10 & Cont. M. Car  & KQ        & Prioritized   & $\rho$-greedy & 500K  & 0.25     & 1.00    & 0.10       & {[}0.8,0.07,1.0{]}  & 67.0 & 0.92 & 85.43                      \\
11 & Cont. M. Car  & Hybrid    & Prioritized   & $\rho$-greedy & 500K  & 0.25     & 1.00    & 0.10       & {[}0.8,0.07,1.0{]}  & 71.22 & 0.76 & 94.72                   \\
12 & Cont. M. Car  & SG        & Prioritized   & $\rho$-greedy & 500K  & 0.25     & 1.00    & 0.10       & {[}0.8,0.07,1.0{]}  & 87.56 & 0.81 & 94.75                    \\
13 & Cont. M. Car  & KQ        & No            & Exploratory       & 100K  & 0.25     & 1.00    & 0.10       & {[}0.8,0.07,1.0{]}  & 36.53 & 0.29 & -21.41                    \\
14 & Cont. M. Car  & KQ        & No            & $\rho$-greedy & 500K  & 0.25     & 1.00    & 0.10       & {[}0.8,0.07,1.0{]}  & 58.42 & 0.21 & 80.92                     \\
15 & Cont. M. Car  & Hybrid    & No            & $\rho$-greedy & 500K  & 0.25     & 1.00    & 0.10       & {[}0.8,0.07,1.0{]}  & {\bf 57.26} & {\bf 0.48} & {\bf 94.96 }                  \\
16 & Cont. M. Car  & SG        & No            & $\rho$-greedy & 500K  & 0.25     & 1.00    & 0.10       & {[}0.8,0.07,1.0{]}  & {\bf 63.13 } & {\bf 0.42 } & {\bf 94.83 } \\      
\hline              
\end{tabular}
\caption{A summary of parameter selection details for our comparison of KQ-Learning(KQ), Hybrid, and Semi-Gradient(SG) methods. In right-most column, we display the limiting model order, training loss (Bellman error) and accumulation of rewards during training. The best results for each problem setting are bolded for emphasis, which ``solve" the problem according to reward benchmarks set by OpenAI. We observe the replay buffer improves learning in the Pendulum domain but yields little benefit in the Mountain Car problem. Interestingly, the Hybrid algorithm in the Pendulum domain attains a smaller training Bellman error but less rewards than the SG approach.}\label{results_table}
\end{table*}

\begin{figure*}
\setcounter{subfigure}{0}
\begin{subfigure}{1.0\columnwidth}
\centering
\includegraphics[width=0.4\linewidth]
                {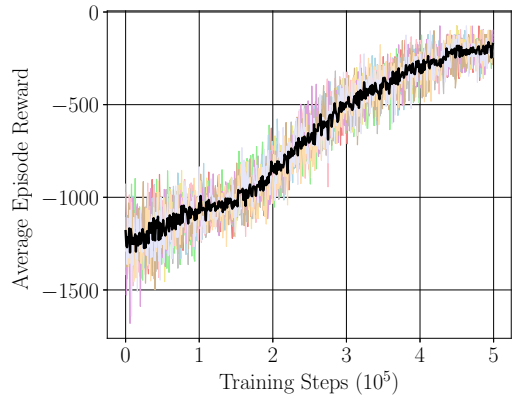}
\caption{Average Training Reward}
\label{subfig:avgreward_pendulum}
\end{subfigure}
\hfill
\begin{subfigure}{1.0\columnwidth}
\centering
\includegraphics[width=0.4\linewidth]
                {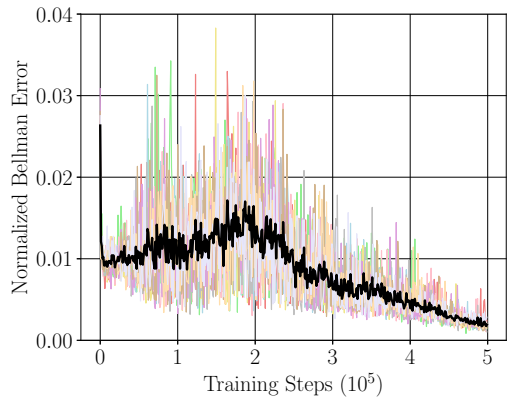}
\caption{Training Bellman Error}
\label{subfig:normbellerr_pendulum}
\end{subfigure}
\hfill
\begin{subfigure}{1.0\columnwidth}
\centering
\includegraphics[width=0.4\linewidth]
                {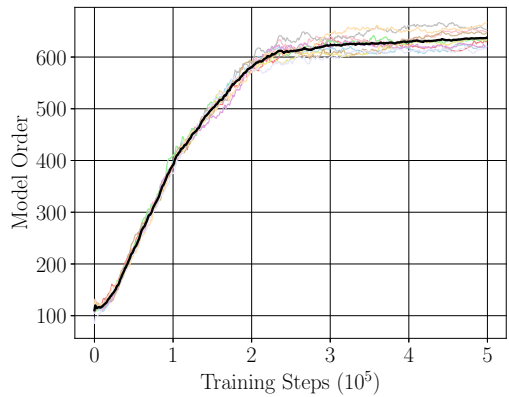}
\caption{Model Order of $Q$}
\label{subfig:modelorder_pendulum}
\end{subfigure}
\caption{
Results of $10$ experiments over $500,000$ training steps were averaged (black curve) to demonstrate the learning progress for the effective, convergent, and parsimonious solution for the Pendulum domain using the Hybrid algorithm with a replay buffer, Row 3 in Table \ref{results_table}. Fig. \ref{subfig:avgreward_pendulum} shows the average reward obtained by the $\rho$-greedy policy during training. Fig. \ref{subfig:normbellerr_pendulum} shows the Bellman error for training samples \eqref{eq:cost_functional} normalized by the Hilbert norm of $Q$, which converges to a small non-zero value. Fig. \ref{subfig:modelorder_pendulum} shows the number of points parameterizing the kernel dictionary of $Q$ during training, which remains under $700$ on average. Overall, we solve Pendulum with a model complexity reduction by orders of magnitude relative to existing methods \cite{mnih2013playing,lillicrap2015continuous}, with a much smaller standard deviation around the average reward accumulation, meaning that these results are replicable.} \label{fig:pendulum}
\end{figure*}


\begin{figure*}

\setcounter{subfigure}{0}
\begin{subfigure}{1.0\columnwidth}
\centering
\includegraphics[width=0.4\linewidth]
                {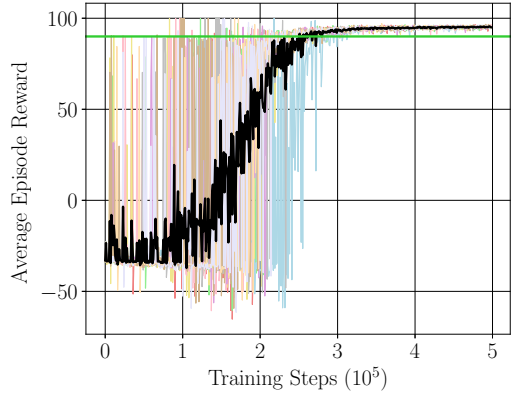}
\caption{Average Training Reward}
\label{subfig:avgreward_car}
\end{subfigure}
\\
\begin{subfigure}{1.0\columnwidth}
\centering
\includegraphics[width=0.4\linewidth]
                {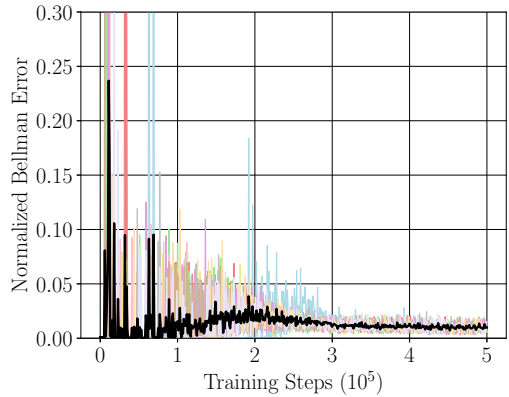}
\caption{Normalized Test Bellman Error}
\label{subfig:normbellerr_car}
\end{subfigure}
\\
\begin{subfigure}{1.0\columnwidth}
\centering
\includegraphics[width=0.4\linewidth]
                {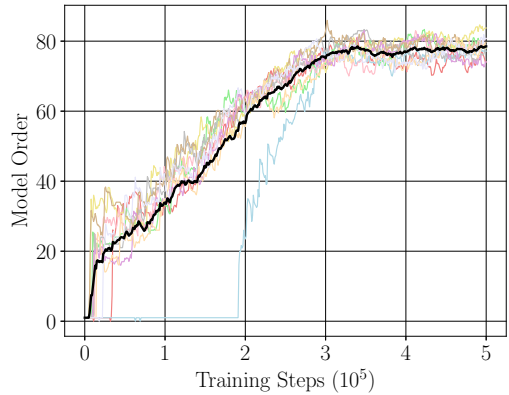}
\caption{Model Order of $Q$}
\label{subfig:modelorder_car}
\end{subfigure}
\caption{
Results of $10$ experiments over $500,000$ training steps were averaged (black curve) to demonstrate the learning progress for Continuous Mountain Car using the Hybrid algorithm with no replay buffer, Row 15 in Table \ref{results_table}. Fig. \ref{subfig:avgreward_car} shows the average reward obtained by the $\rho$-greedy policy during training. An average reward over 90 (green) indicates that we have solved Continuous Mountain Car, steering towards the goal location. Fig. \ref{subfig:normbellerr_car} shows the normalized Bellman error during training, which converges to a small non-zero value. Fig. \ref{subfig:modelorder_car} shows the number of points parameterizing the kernel dictionary of $Q$ during training, which remains under $80$ on average. Overall, we solve Continuous Mountain Car with a complexity reduction by orders of magnitude relative to existing methods\cite{mnih2013playing,lillicrap2015continuous}. We observe that learning progress has higher variance, which we hypothesize is related to the sparsity of the reward signal. \vspace{-4.5mm}} \label{fig:mountain_car}
\end{figure*}
\begin{figure*} \label{fig:vis_figure}
\setcounter{subfigure}{0}  
\begin{subfigure}{1.0\columnwidth}
\centering
\includegraphics[width=0.6\linewidth]
                {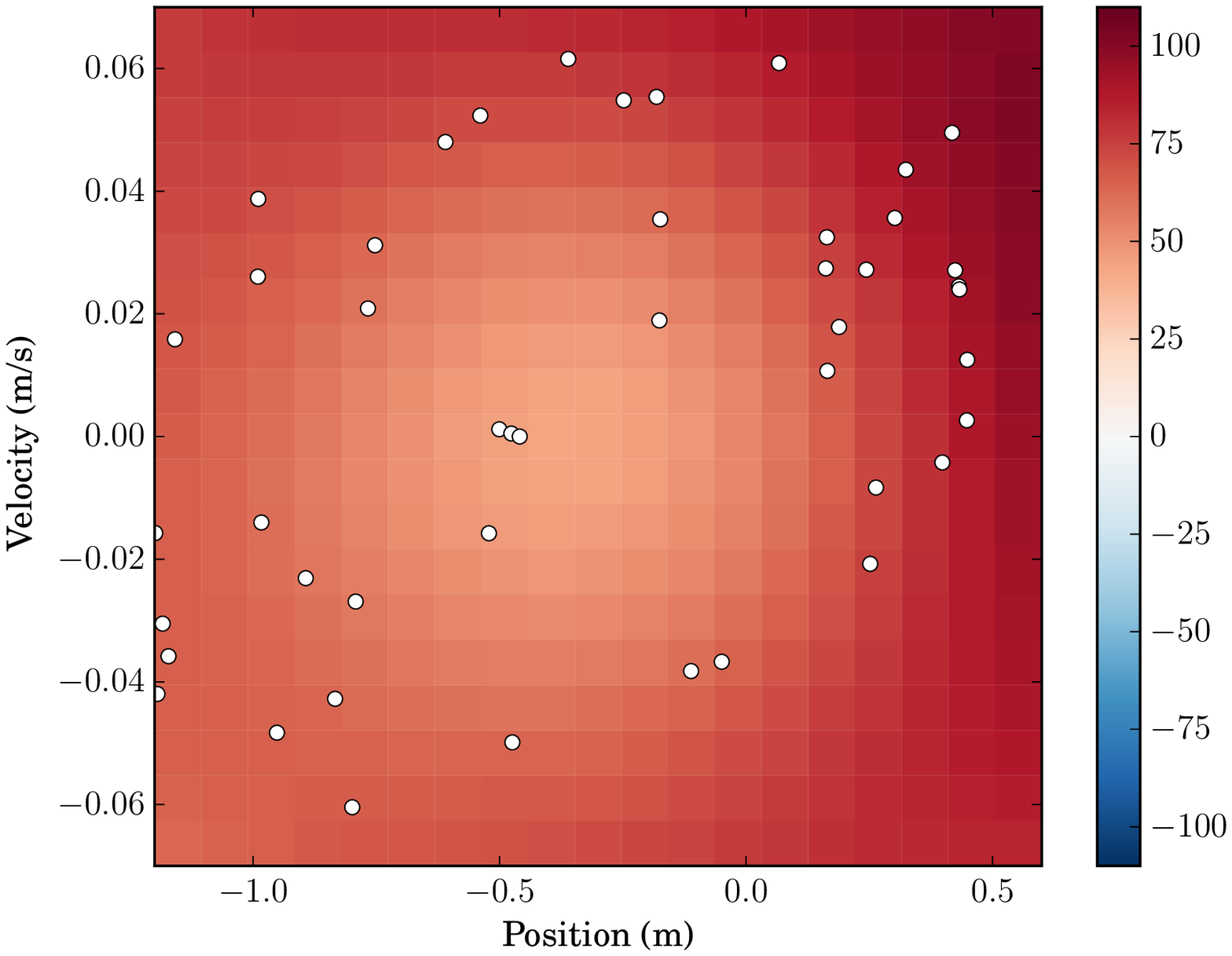}
\caption{ Value function $V(\bbs)$ derived from limiting $Q(\bbs,\bba)$ }
\label{subfig:value}
\end{subfigure}
\\
\begin{subfigure}{1.0\columnwidth}
\centering
\includegraphics[width=0.6\linewidth]
                {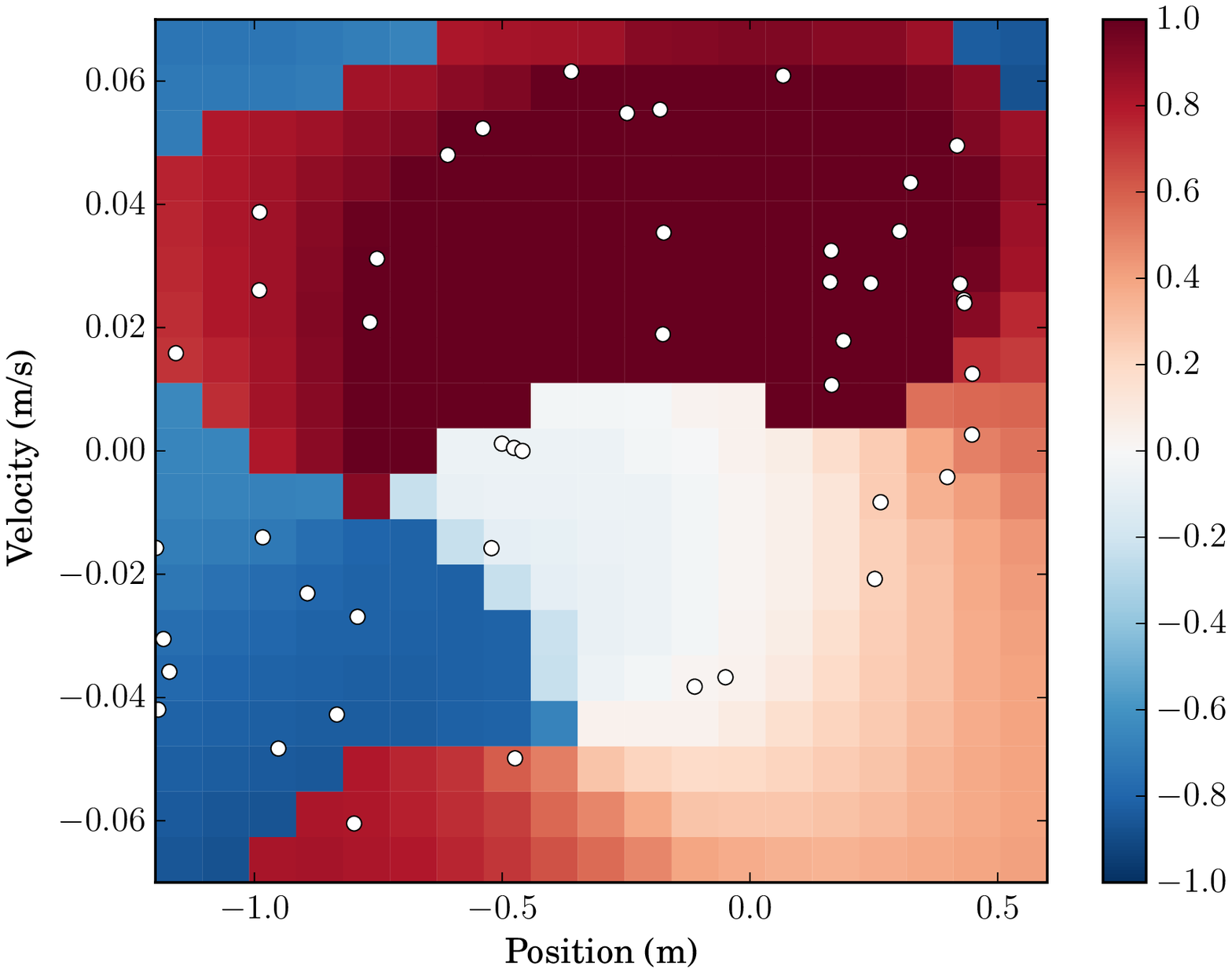}
\caption{Policy  $\pi(\bbs)$ derived from limiting $Q(\bbs,\bba)$ }
\label{subfig:policy}
\end{subfigure}
\caption{For the mountain car problem, the learned $Q$-function is easily interpretable: we may visualize the value function, $V(s) = \max_a Q(s,a)$ (\ref{subfig:value}) and corresponding policy $\pi(s) = \argmax_a Q(s,a)$ (\ref{subfig:policy}). In Fig. \ref{subfig:value}, the color indicates the value of the state, which is highest (dark red) near the goal $0.6$. At this position, for any velocity, the agent receives an award of 100 and concludes the episode. In Fig. \ref{subfig:policy}, the color indicates the force on the car (action), for a given position and velocity (state). The learned policy takes advantage of the structure of the environment to accelerate the car without excess force inputs. The dictionary points are pictured in white and provide coverage of the state-action space.  
\vspace{-5mm}
}
\end{figure*}

\vspace{-4mm}
\section{Experiments}\label{sec:experiments}

We shift focus to experimentation of the methods developed and analyzed in the previous sections. Specifically, we benchmark the proposed algorithms on two classic control problems, the Inverted Pendulum \cite{yoshida1999swing,argyriou2009there} and the Continuous Mountain Car \cite{Moore90efficientmemory}, which are featured in OpenAI Gym \cite{openaigym}. 

In the Inverted Pendulum problem, the state space is $p=3$ dimensional, consisting of the sine of the angle of the pendulum, the cosine of the angle, and the angular velocity, bounded within $\lbrack -1.0,1.0 \rbrack$,  $\lbrack -1.0,1.0 \rbrack$, and $\lbrack -8.0,8.0 \rbrack$ respectively. The action space is $q=1$ dimensional: joint effort, within the interval $\lbrack -2.0,2.0 \rbrack$. The reward function is
$r(\theta, \dot{\theta}, a) = -(\theta^2 + 0.1 \dot{\theta}^2 + 0.001 a^2) $, where $\theta$ is the angle of the pendulum relative to vertical, and $\dot{\theta}$ is the angular velocity. The goal of the problem is to balance the pendulum at the unstable equilibrium where $\theta=0$.

In the Continuous Mountain Car problem, the state space is $p=2$ dimensional, consisting of position and velocity, bounded within $\lbrack -1.2, 0.6 \rbrack$ and $\lbrack -0.07, 0.07 \rbrack$, respectively. The action space is $q=1$ dimensional: force on the car, within the interval $\lbrack -1,1 \rbrack$. The reward function is 100 when the car reaches the goal at position $0.45$, and $-0.1a^2$ for any action $a$. For each training episode, the start position of the car was initialized uniformly at random in the range $\lbrack -0.6, 0.4 \rbrack$.

For all experiments with the Inverted Pendulum and the Continuous Mountain Car problems, we used Gaussian kernels with a fixed non-isotropic bandwidth. The relevant parameters are the step-sizes $\alpha$ and $\beta$, the regularizer $\lambda$, and the approximation error constant, $C$, where we fix the compression budget $\eps=C \alpha^2$. These learning parameters were tuned through a grid search procedure, and are summarized in Table \ref{results_table}.

We investigate two methods for exploration as the agent traverses the environment. When using an exploratory policy, actions are selected uniformly at random from the action space. 
When using an $\rho$-greedy policy, we select actions randomly with probability $1$, which linearly decays to $0.1$ after $10^5$ exploratory training steps.
In addition, we explore the use of a replay buffer. This method re-reveals past data to the agent uniformly at random. For the Mountain Car problem, we also use prioritized memory which replays samples based on the magnitude of their temporal action difference.

A comprehensive summary of our experimental results may be found in Table \ref{results_table}. We bold which methods perform best across many different experimental settings. Interestingly, playback buffers play a role in improving policy learning in the Pendulum domain but not for Mountain Car, suggesting that their merit demands on the reward structure of the MDP.

 We spotlight the results of this experiment in the Pendulum domain for the Hybrid algorithm in Figure \ref{fig:pendulum}: here we plot the normalized Bellman test error Fig. \ref{subfig:normbellerr_pendulum}, defined by the sample average approximation of \eqref{eq:cost_functional} divided by the Hilbert norm of $Q_t$ over a collection of generated test trajectories, as well as the average rewards  during training (Fig. \ref{subfig:avgreward_pendulum}), and the model order, i.e., the number of training examples in the kernel dictionary (Fig. \ref{subfig:modelorder_pendulum}), all relative to the number of training samples processed.

Observe that the Bellman test error converges and the interval average rewards approach $-200$, which is comparable to top entries on the  OpenAI Leaderboard \cite{openaigym}, such as Deep Deterministic Policy Gradient \cite{lillicrap2015continuous} . Moreover, we obtain this result with a complexity reduction by orders of magnitude relative to existing methods for $Q$-function and policy representation. 
 This trend is corroborated for the Continuous Mountain Car in Figure \ref{fig:mountain_car}: the normalized Bellman error converges and the model complexity remains moderate.  Also, observe that the interval average rewards approach $90$, which is the benchmark used to designate a policy as ``solving" Continuous Mountain Car.
 
 Additionally, few heuristics are required to ensure $KQ$-Learning converges in contrast to neural network approaches to $Q$-learning. One shortcoming of our implementation is its sample efficiency, which could improved through a mini-batch approach. Alternatively, variance reduction, acceleration, or Quasi-Newton methods would improve the learning rate.

A feature of our method is the interpretability of the resulting $Q$ function, which we use to plot the value function (\ref{subfig:value}) and policy (\ref{subfig:policy}). One key metric is the coverage of the kernel points in the state-action space. We can make conclusions about the importance of certain parts of the space for obtaining as much value as possible by the density of the model points throughout the space. This may have particular importance in mechanical or econometric applications, where the model points represent physical phenomena or specific events in financial markets. 

\section{Conclusion} \label{sec:conclusion}
In this paper, we extended the nonparametric optimization approaches in \cite{pkgtd} from policy evaluation to policy learning in continuous Markov Decision Problems. In particular, we reformulated the task of policy learning defined by the Bellman optimality equation as a non-convex function-valued stochastic program with nested expectations. We hypothesize that the Bellman fixed point belongs to a reproducing Kernel Hilbert Space, motivated by their efficient semi-parametric form. By applying functional stochastic quasi-gradient method operating in tandem with greedily constructed subspace projections, we derived a new efficient variant of $Q$ learning which is guaranteed to converge almost surely in continuous spaces, one of the first results of this type.  

Unlike the policy evaluation setting, in policy learning we are forced to confront fundamental limitations associated with non-convexity and the explore-exploit tradeoff. To do so, we adopt a hybrid policy learning situation in which some actions are chosen greedily and some are chosen randomly. Through careful tuning of the proportion of actions that are greedy versus exploratory, we are able to design a variant of $Q$ learning which learns good policies on some benchmark tasks, namely, the Continuous Mountain Car and the Inverted Pendulum, with orders of magnitude fewer training examples than existing approaches based on deep learning. Further, owing to the kernel parameterization of our learned $Q$ functions, they are directly interpretable: the training points which are most vital for representing the minimal Bellman error action-value function are retained and automatically define its feature representation.

\bibliographystyle{IEEEtran}
\bibliography{IEEEfull,bibliography} 
\vspace{-1.7cm}
\begin{biography}[{\includegraphics[width=25mm,height=32mm,clip,keepaspectratio]{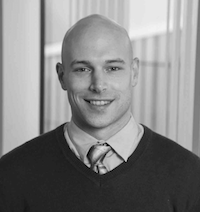}}]
{Alec Koppel} began as a Research Scientist at the U.S. Army Research Laboratory in the Computational and Information Sciences Directorate in September of 2017. He completed his Master's degree in Statistics and Doctorate in Electrical and Systems Engineering, both at the University of Pennsylvania (Penn) in August of 2017. He is also a participant in the Science, Mathematics, and Research for Transformation (SMART) Scholarship Program sponsored by the American Society of Engineering Education. Before coming to Penn, he completed his Master's degree in Systems Science and Mathematics and Bachelor's Degree in Mathematics, both at Washington University in St. Louis (WashU), Missouri. His research interests are in the areas of signal processing, optimization and learning theory. His current work focuses on optimization and learning methods for streaming data applications, with an emphasis on problems arising in autonomous systems. He co-authored a paper selected as a Best Paper Finalist at the 2017 IEEE Asilomar Conference on Signals, Systems, and Computers.
\end{biography} 

\vspace{-1.5cm}
\begin{biography}[{\includegraphics[width=25mm,height=32mm,clip,trim={0 40px 0 0 },clip,keepaspectratio]{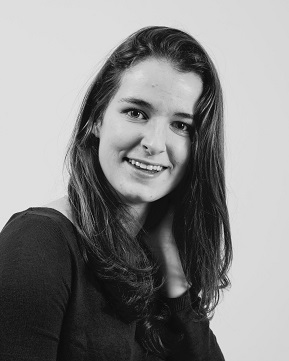}}]
{Ekaterina Tolstaya} is a doctoral student in the Department of Electrical and Systems Engineering, University of Pennsylvania, Philadelphia, PA, and a National Science Foundation Graduate Research Fellow. She received B.Sc. degrees  in Electrical Engineering and Computer Science from University of Maryland, College Park, MD, in 2016, and the M.S.E. degree in Robotics from University of
Pennsylvania, Philadelphia, PA, in 2017. Her research interests include reinforcement learning, aerial robotics, and multi-agent systems. 
\end{biography} 

\vspace{-1.5cm}
\begin{biography}[{\includegraphics[width=25mm,height=32mm,clip,keepaspectratio]{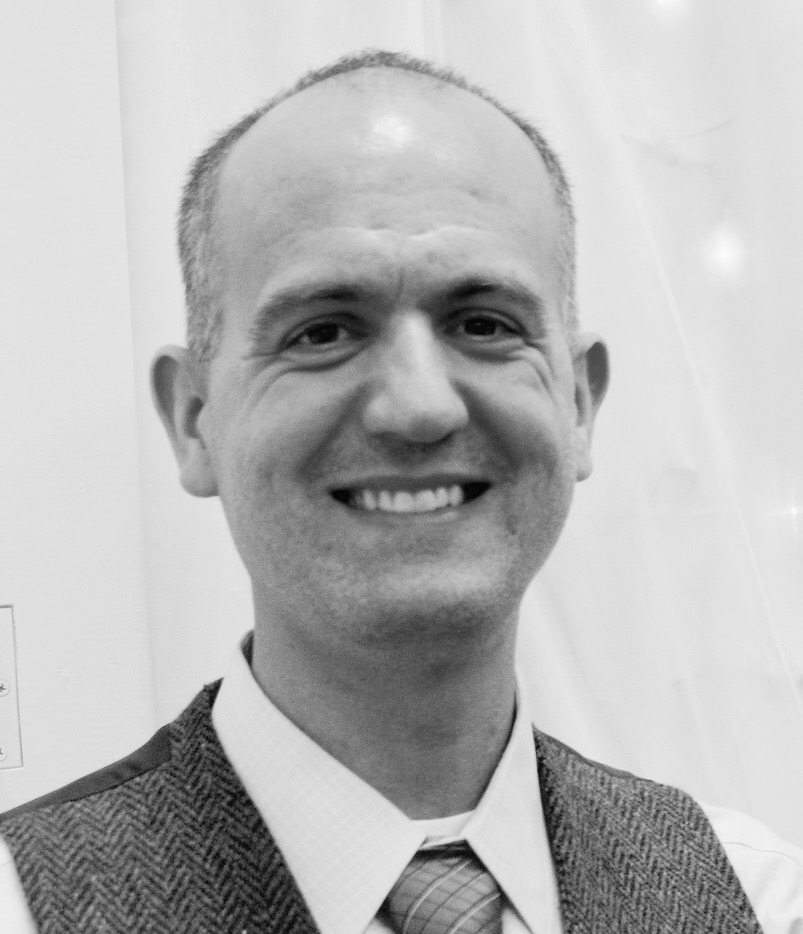}}]
{Ethan A. Stump}  received the B.S. degree from the Arizona State University, Tempe, and the M.S. and PhD degrees from the University of Pennsylvania, Philadelphia, all in mechanical engineering. He is a researcher within the U.S. Army Research Laboratory’s Computational and Information Sciences Directorate, where he has worked on developing mapping and navigation technologies to enable baseline autonomous capabilities for teams of ground robots and on developing controller synthesis for managing the deployment of multi-robot teams to perform repeating tasks such as persistent surveillance by tying them formal task specifications.
\end{biography} 

\vspace{-.8cm}
\begin{biography}[{\includegraphics[width=25mm,height=32mm,clip,keepaspectratio]{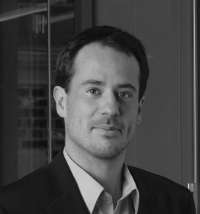}}]
{Alejandro Ribeiro}
received the B.Sc. degree in electrical
engineering from the Universidad de la Republica
Oriental del Uruguay, Montevideo, in 1998 and the
M.Sc. and Ph.D. degree in electrical engineering from
the Department of Electrical and Computer Engineering,
the University of Minnesota, Minneapolis in 2005
and 2007. From 1998 to 2003, he was a member
of the technical staff at Bellsouth Montevideo. After
his M.Sc. and Ph.D studies, in 2008 he joined the
University of Pennsylvania (Penn), Philadelphia, where
he is currently the Rosenbluth Associate Professor at
the Department of Electrical and Systems Engineering. His research interests
are in the applications of statistical signal processing to the study of networks
and networked phenomena. His current research focuses on wireless networks,
network optimization, learning in networks, networked control, robot teams, and
structured representations of networked data structures. Dr. Ribeiro received the
2012 S. Reid Warren, Jr. Award presented by Penn's undergraduate student
body for outstanding teaching, the NSF CAREER Award in 2010, and student
paper awards at the 2013 American Control Conference (as adviser), as well as
the 2005 and 2006 International Conferences on Acoustics, Speech and Signal
Processing. Dr. Ribeiro is a Fulbright scholar and a Penn Fellow.
\end{biography} 

 \section{Appendices}\label{sec:appendix}
 
\subsection*{Appendix A: Proof of Auxiliary Results}\label{sec:apx_a}

We turn to establishing some technical results which are necessary precursors to the proofs of the main stability results. 
\begin{proposition}\label{prop_proj_grad_err} Given independent identical realizations $(\mathbf{s}_t,\mathbf{a}_t,\mathbf{s}_t')$ of the random triple $(\mathbf{s},\mathbf{a},\mathbf{s}')$, the difference between the projected stochastic functional quasi-gradient and the stochastic functional quasi-gradient of the instantaneous cost is bounded for all $t$ as
\begin{align} \label{eq:prop1proof}
\!\!\!\!   \|\! \tilde{\nabla}_{\!Q} J(\!Q_t, \! z_{t\!+\!1} \!;  \mathbf{s}_t\!, \! \mathbf{a}_t\!, \! \mathbf{s}_t'\!)\! -\!\!  \hat{\nabla}_{\!Q} J(\!Q_t, \! z_{t\!+\!1} \!;  \mathbf{s}_t\!, \! \mathbf{a}_t\!, \!\mathbf{s}_t'\!) \!\|_{\ccalH} \!\!\leq \!\!\! \frac{\epsilon_t}{\alpha_t}
\end{align}
Where $\alpha_t > 0$ denotes the algorithm step size and $\epsilon_t > 0$ is the compression budget parameter of the KOMP algorithm. 
\end{proposition}

\begin{myproof}
As in Proposition 1 of \cite{pkgtd}, Consider the square-Hilbert norm difference of $\tilde{\nabla}_Q J(Q_t, z_{t+1};   \mathbf{s}_t,\mathbf{a}_t, \mathbf{s}_t')$ and $\hat{\nabla}_Q  J(Q_t, z_{t+1};   \mathbf{s}_t,\mathbf{a}_t, \mathbf{s}_t')$ defined by \eqref{eq:unprojectedgradproof} and \eqref{eq:projectedgradproof}
\begin{align} \label{eq:prop1aproof}
\| \tilde{\nabla}_Q  J(Q_t, z_{t+1};   \mathbf{s}_t,\mathbf{a}_t, \mathbf{s}_t') - \hat{\nabla}_Q J(Q_t, z_{t+1}; \mathbf{s}_t,\mathbf{a}_t,\mathbf{s}_t') \|_{\ccalH} & =
 \\  
 \| & (Q_t -  \mathcal{P}_{\ccalH_{\mathbf{D}_{t+1}}}  \lbrack Q_t - \alpha_t \hat{\nabla}_Q J(Q_t, z_{t+1}; \mathbf{s}_t, \mathbf{a}_t, \mathbf{s}'_t) \rbrack  )/\alpha_t 
  -   \hat{\nabla}_Q J(Q_t, z_{t+1}\!; \mathbf{s}_t, \mathbf{a}_t, \mathbf{s}'_t) \|_\ccalH^2 \nonumber
\end{align}
Multiply and divide $\hat{\nabla}_Q  J(Q_t, z_{t+1};   \mathbf{s}_t,\mathbf{a}_t, \mathbf{s}_t')$ by $\alpha_t$ and reorder terms to write
\begin{align} \label{eq:prop1bproof}
 & \|  \frac{(Q_t -  \alpha_t \hat{\nabla}_Q J(Q_t, z_{t+1}\!; \mathbf{s}_t, \mathbf{a}_t, \mathbf{s}'_t)  )}{\alpha_t} 
- \frac{(\mathcal{P}_{\ccalH_{\mathbf{D}_{t+1}}}  \lbrack Q_t - \alpha_t \hat{\nabla}_Q J(Q_t, z_{t+1}; \mathbf{s}_t, \mathbf{a}_t, \mathbf{s}'_t) \rbrack  )}{\alpha_t} 
\|_\ccalH^2   
\nonumber \\ 
& =  \frac{1}{\alpha_t^2} \| (Q_t -  \alpha_t \hat{\nabla}_Q J(Q_t, z_{t+1}\!; \mathbf{s}_t, \mathbf{a}_t, \mathbf{s}'_t)  )
- (\mathcal{P}_{\ccalH_{\mathbf{D}_{t+1}}}  \lbrack Q_t - \alpha_t \hat{\nabla}_Q J(Q_t, z_{t+1}; \mathbf{s}_t, \mathbf{a}_t, \mathbf{s}'_t) \rbrack  ) \|_\ccalH^2
\nonumber \\ 
& =  \frac{1}{\alpha_t^2} \| \tilde{Q}_{t+1} - Q_{t+1} \|_\ccalH^2 \leq \frac{\epsilon_t^2}{\alpha_t^2}
\end{align}
where we have pulled the nonnegative scalar $\alpha_t$ outside of the norm on the second line and substituted the definition of $\tilde{Q}_{t+1}$ and $Q_{t+1}$. We also apply the KOMP residual stopping criterion from Algorithm 2, $\| \tilde{Q}_{t+1} - Q_{t+1} \| \leq \epsilon_t$ to yield \eqref{eq:prop1proof}.
\end{myproof}

\begin{lemma}\label{main_lemma} Denote the filtration $\mathcal{F}_t$ as the time-dependent sigma-algebra containing the algorithm history $( \lbrace Q_u, z_u\rbrace_{u=0}^t   \cup \lbrace \mathbf{s}_u, \mathbf{a}_u, \mathbf{s}_u' \rbrace_{u=0}^{t-1}) \subset \mathcal{F}_t$. Let Assumptions \ref{as:compact}-\ref{as:mean_variance_q} hold true and consider the sequence of iterates defined by Algorithm \ref{qlearning}. Then:
\begin{enumerate}[i.]
\item The conditional expectation of the Hilbert-norm difference  of action-value functions at the next and current iteration satisfies the relationship \label{main_lemma_1}
\begin{equation} \label{eq:lemma1iproof}
 \! \! \mathbb{E} \lbrack \| Q_{t+1} \!- \!Q_t \|^2_{\ccalH}\!  \! \mid\!\! \mathcal{F}_t \rbrack \!\! \leq \! 2 \alpha^2_t \!(\!G_{\delta}^2G_Q^2 \!+ \! \! \lambda \! D^2 \!) \! + \! 2  \epsilon_t^2
\end{equation}
\item The auxiliary sequence $z_t$ with respect to the conditional expectation of the temporal action difference $\bar{\delta}_t$ (defined in Assumption \ref{as:mean_variance_delta}) satisfies \label{main_lemma_2}
\begin{align} \label{eq:lemma1iiproof}
\!\mathbb{E} \lbrack  (z_{t+1} -  \bar{\delta_t})^2 \mid \! \mathcal{F}_t\!  \rbrack & \leq (1-\beta_t)(z_t - \bar{\delta}_{t-1})^2  + \frac{L_Q}{\beta_t} \|Q_t \!-\! Q_{t-1}\|^2_\ccalH + 2 \beta_t^2 \sigma_{\delta}^2
\end{align}
\item Algorithm \ref{qlearning} generates a sequence of Q-functions that satisfy the stochastic descent property with respect to the Bellman error $J(Q)$ [cf. \eqref{eq:lossprob}]: \label{main_lemma_3}
\begin{align} \label{eq:lemma1iiiproof}
\mathbb{E}[J(\!Q_{t+1}\!)\given\ccalF_t] \leq   J(\!Q_t\!)
- \alpha_t \!\!\left(\!\!1\! -\! \frac{\alpha_t G_{Q}^2 }{ \beta_t}\!\!\right) \|\nabla_Q J(Q)\|^2\! \!
\quad+\!  \frac{\beta_t}{2} \mathbb{E} \lbrack ( \bar{\delta_t} - z_{t+1} )^2 \mid \mathcal{F}_t \rbrack+\frac{L_Q \sigma_Q^2 \alpha_t^2}{2}  + \eps_t \|\nabla_Q J(Q_t) \|_{\ccalH} \; ,
\end{align}
\end{enumerate}
\end{lemma}

 \begin{myproof} \textit{Lemma \ref{main_lemma}\eqref{main_lemma_1} }
Consider the Hilbert-norm difference of action-value functions at the next and current iteration and use the definition of $Q_{t+1}$
\begin{equation} \label{eq:aproof}
\!\! \|Q_{t+1} \!\! - \!\! Q_t \! \|_\ccalH^2 \!\! = \!\! \alpha_t^2 \| \!\tilde{\nabla}_Q J(Q_t, z_{t+1}\!;\! (\mathbf{s}_t,\mathbf{a}_t),(\mathbf{s}_t'\!,\mathbf{a}_t'\!) \|_\ccalH^2 \!
\end{equation}
We add and subtract the functional stochastic quasi-gradient $\hat{\nabla}_Q J(Q_t, z_{t+1}; (\mathbf{s}_t,\mathbf{a}_t),(\mathbf{s}_t',\mathbf{a}_t')$ from \eqref{eq:aproof} and apply the triangle inequality $(a+b)^2 \leq 2a^2 + 2b^2$ which holds for any $a,b>0$.
\begin{align} \label{eq:bproof}
\!\!\|Q_{t+1} \! - \! Q_t\|_\ccalH^2  
&\leq 2 \alpha_t^2 \|\hat{\nabla}_Q J(Q_t, z_{t+1}; (\mathbf{s}_t,\mathbf{a}_t),(\mathbf{s}_t',\mathbf{a}_t') \|_\ccalH^2 
 + \! 2 \alpha_t^2  \|\hat{\nabla}_Q J(Q_t, z_{t+1}; (\mathbf{s}_t,\mathbf{a}_t),(\mathbf{s}_t',\mathbf{a}_t')  
  - \!\tilde{\nabla}_Q J(Q_t, z_{t+1}\!; (\mathbf{s}_t\!,\mathbf{a}_t\!)\!,\!(\mathbf{s}_t'\!,\mathbf{a}_t') \|_\ccalH^2
\end{align}
 Now, we may apply Proposition 1 to the second term. Doing so and computing the expectation conditional on the filtration $\mathcal{F}_t$ yields
 \begin{align} \label{eq:cproof}
 \!\!   \mathbb{E}  &\lbrack \|Q_{t+1} - Q_t\|_\ccalH^2 \rbrack \mid \mathcal{F} \rbrack  
   2 \alpha_t^2 \mathbb{E} \lbrack \|\hat{\nabla}_Q J(Q_t, z_{t+1}; (\mathbf{s}_t,\mathbf{a}_t),(\mathbf{s}_t',\mathbf{a}_t') \|_\ccalH^2 \mid \mathcal{F}_t \rbrack \! + \! 2 \epsilon_t^2
 \end{align}
Using the Cauchy-Schwarz inequality together with the Law of Total Expectation and the definition of the functional stochastic quasi-gradient to upper estimate the first term on the right-hand side of \eqref{eq:cproof} as
\begin{align} \label{eq:dproof}
 \mathbb{E}  \lbrack  \| Q_{t+1}  &-  Q_t\|^2_\ccalH \mid \mathcal{F}_t \rbrack  
 \leq  2 \alpha_t^2 \mathbb{E} \lbrace \| \gamma \kappa((\mathbf{s}_t',\mathbf{a}_t'),\cdot) - \! \kappa((\mathbf{s}_t,\mathbf{a}_t),\!\cdot\!) \|_\ccalH^2 
\times \mathbb{E} \lbrack z_{t+1}^2 \! \mid \mathbf{s}_t, \! a_t \rbrack \mid \mathcal{F}_t \! \rbrace 
\!\! + \!\! 2 \alpha_t^2 \!\! \lambda \|Q_t\|_\ccalH^2 \! \!\! + \!\! 2 \epsilon_t^2
\end{align}
Now, use the fact that $z_{t+1}$ has a finite second conditional moment [cf. \eqref{eq:mean_variance_delta}], yielding
\begin {align}  \label{eq:eproof}
\mathbb{E}  \lbrack & \|Q_{t+1} -  Q_t\|^2_\ccalH \mid \mathcal{F}_t \rbrack
\leq  2 \alpha_t^2 G_\delta^2 \mathbb{E} \lbrack \| \gamma \kappa((\mathbf{s}_t',\mathbf{a}_t'),\cdot) 
- \kappa((\mathbf{s}_t,\mathbf{a}_t),\cdot) \|_\ccalH^2  \mid \mathcal{F}_t \rbrack 
+ 2 \alpha_t^2 \lambda \|Q_t\|_\ccalH^2 + 2 \epsilon_t^2
\end{align}
From here, we may use the fact that the functional gradient of the temporal action-difference $\gamma \kappa((\mathbf{s}_t',\mathbf{a}_t'),\cdot) - \kappa((\mathbf{s}_t,\mathbf{a}_t),\cdot)$ has a finite second conditional moment \eqref{eq:mean_variance_delta} and that the $Q$ function sequence is bounded \eqref{eq:compactnessproof} to write:
\begin{equation} \label{eq:fproof}
\mathbb{E} \lbrack \|Q_{t+1} \!\! - \! Q_t\|^2_\ccalH \mid \mathcal{F}_t \rbrack \leq 2 \alpha_t^2 (G_\delta^2 G_V^2  \! +  \! \! \lambda^2 D^2) \! + \! 2 \epsilon_t^2
\end{equation}
which is as stated in Lemma \ref{main_lemma}\eqref{main_lemma_1}.
\end{myproof}

 \begin{myproof} \textit{Lemma \ref{main_lemma}\eqref{main_lemma_2} }
Begin by defining the scalar quantity $e_t$ as the difference of mean temporal-action differences scaled by the forgetting factor $\beta_t$, i.e. $e_t = (1-\beta_t)(\bar{\delta_t} - \bar{\delta}_{t-1})$. Then, we consider the difference of the evolution of the auxiliary variable  $z_{t+1}$ with respect to the conditional mean temporal action difference $\bar{\delta}_t$, plus the difference of the mean temporal differences:
\begin{align} \label{eq:gproof}
z_{t+1} - \bar{\delta}_t + e_t =  & (1-\beta_t)z_t + \beta_t\delta_t - \lbrack(1-\beta_t)\bar{\delta_t} + \beta_t \bar{\delta_t} \rbrack
 + (1-\beta_t)(\bar{\delta_t} - \bar{\delta}_{t-1}) 
\end{align}
where we make use of the definition of $z_{t+1}$, the fact that $\bar{\delta}_t = \lbrace (1-\beta_t)\bar{\delta}_t + \beta_t \bar{\delta}_t \rbrace$ and the definition of $e_t$ on the right-hand side of \eqref{eq:gproof}. Observe that the result then simplifies to $z_{t+1} - \bar{\delta}_t + e_t=(1-\beta_t)z_t +  \beta_t(\bar{\delta_t} - \bar{\delta}_{t-1}) $ by grouping like terms and canceling the redundant $\bar{\delta}_t$. Squaring \eqref{eq:gproof}, using this simplification, yields
\begin{align} \label{eq:hproof}
(z_{t+1} - &\bar{\delta}_t + e_t)^2 
=  (1-\beta_t)^2(z_t - \bar{\delta}_{t-1})^2 
+ \beta_t^2(\delta_t - \bar{\delta}_t)^2 
+ 2(1-\beta_t)\beta_t(z_t-\bar{\delta}_{t-1})(\delta_t - \bar{\delta_t}) 
\end{align}
Now, we compute the expectation conditioned on the algorithm history $\mathcal{F}_t$ to write
\begin{align} \label{eq:iproof}
\mathbb{E}\lbrack(z_{t+1} & - \bar{\delta}_t + e_t)^2 \mid \mathcal{F}_t\rbrack 
=  (1-\beta_t)^2(z_t - \bar{\delta}_{t-1})^2 + \beta_t^2  \mathbb{E}  \lbrack(\delta_t - \bar{\delta}_t)^2 \mid \mathcal{F}_t\rbrack  
+ 2(1-\beta_t)\beta_t(z_t-\bar{\delta}_{t-1})\mathbb{E}\lbrack(\delta_t - \bar{\delta_t}) \mid \mathcal{F}_t\rbrack
\end{align}
We apply the assumption that the temporal action difference $\delta_t$ is an unbiased estimator for its conditional mean $\bar{\delta}_t$ with finite variance (Assumption \ref{as:mean_variance_delta}) to write
\begin{equation} \label{eq:kproof}
\!\! \mathbb{E} \lbrack (z_{t+1} \!\! - \! \bar{\delta}_t \! +  \! e_t) \! \given  \! \ccalF_t \rbrack \! = \! (1 \!- \! \beta_t)^2 (z_t \! - \! \bar{\delta}_{t-1})^2 \! + \! \beta_t^2 \! \sigma_\delta^2
\end{equation}
We obtain an upper estimate on the conditional mean square of $z_{t+1} - \bar{\delta}_t$ by using the inequality $\| a + b \|^2 \leq (1+\rho) \| a \|^2 + (1+1 \rho) \| b \|^2$ which holds for any $\rho > 0$: set $a=z_{t+1} - \bar{\delta}_t + e_t$, $b = -e_t$, $\rho = \beta_t$ to write
\begin{equation} \label{eq:lproof}
\!\!\!\!\!(z_{t+1} \! - \! \bar{\delta}_t)^2 \! \leq \! (1+\beta_t)(z_{t+1} \! - \! \bar{\delta}_t \! + \! e_t)^2 \! + \! \left(1 \! + \! \frac{1}{\beta_t}\right) e_t^2
\end{equation}
Observe that \eqref{eq:lproof} provides an upper-estimate of the square sub-optimality $(z_{t+1} - \bar{\delta}_t)^2$ in terms of the squared error sequence $(z_{t+1} \! - \! \bar{\delta}_t \! + \! e_t)^2 $. Therefore, we can compute the expectation of \eqref{eq:lproof} conditional on $\ccalF_t$ and substitute \eqref{eq:kproof} for the terms involving the error sequence $(z_{t+1} \! - \! \bar{\delta}_t \! + \! e_t)^2 $, which results in gaining a factor of $(1+\beta_t)$ on the right-hand side. Collecting terms yields
\begin{align} \label{eq:mproof}
\mathbb{E}  \lbrack (z_{t+1} - \bar{\delta}_t)^2 \mid \mathcal{F}_t \rbrack  
  = (1+\beta_t)  \lbrack (1-\beta_t)^2(z_t\! -\! \bar{\delta}_{t-1})^2 + \beta_t^2\sigma_\delta^2 \rbrack + \left(\!\!\frac{1+\beta_t}{\beta_t}\!\!\right) e_t^2 
\end{align} 
Using the fact that $(1-\beta_t^2)(1-\beta_t) \leq (1-\beta_t)$ for the first term and $(1-\beta_t)\beta_t^2 \leq 2 \beta_t^2$ for the second to simplify
\begin{align} \label{eq:nproof}
\mathbb{E} \lbrack (z_{t+1} - \bar{\delta}_t)^2 \mid \mathcal{F}_t \rbrack  
=  & (1-\beta_t)(z_t - \bar{\delta}_{t-1})^2 + 2\beta_t^2\sigma_\delta^2  
 + \left(\frac{1+\beta_t}{\beta_t}\right) e_t^2
\end{align}
We can bound the term involving $e_t$, which represents the difference of mean temporal differences. By definition, we have $|e_t| = (1- \beta_t)|(\bar{\delta}_t  -  \bar{\delta}_{t-1})|$:
\begin{equation} \label{eq:oproof}
(1\!-\! \beta_t)|(\bar{\delta}_t \! - \! \bar{\delta}_{t-1})| \! \leq \! (1 \!- \! \beta_t) L_Q \| Q_t \! - \! Q_{t-1} \! \|_\ccalH \; ,
\end{equation}
where we apply the Lipschitz continuity of the temporal difference with respect to the action-value function [cf. \eqref{eq:deltaqproof}]. Substitute the right-hand side of \eqref{eq:oproof} and simplify the expression in the last term as $(1-\beta_t^2) / \beta_t \leq 1/\beta_t$ to conclude \eqref{eq:lemma1iiiproof}.
 \end{myproof} 
 
 \begin{myproof} \textit{Lemma \ref{main_lemma}\eqref{main_lemma_3} }
Following the proof of Theorem 4 of \cite{wang2017stochastic}, we begin by considering the Taylor expansion of $J(Q)$ and applying the fact that it has Lipschitz continuous functional gradients to upper-bound the second-order terms. Doing so yields the quadratic upper bound:
\begin{align} \label{eq:pproof}
J(Q_{t+1}) &\leq   J(Q_t) + \langle \nabla J(Q_t) , Q_{t+1} - Q_t \rangle_\ccalH 
 + \frac{L_Q}{2} \| Q_{t+1} - Q_t \|_\ccalH^2 \; .
\end{align}
Substitute the fact that the difference between consecutive action-value functions is the projected quasi-stochastic gradient $Q_{t+1} - Q_t  = -\alpha_t \tilde{\nabla}_Q J(Q_t, z_{t+1}; \mathbf{s}_t,\mathbf{a}_t,\mathbf{s}_t')$ \eqref{eq:qprojectedupdateproof} into \eqref{eq:pproof} . 
\begin{align} \label{eq:paroof}
J(Q_{t+1}) \leq  & J(Q_t) - \alpha_t \langle \nabla J(Q_t) , \tilde{\nabla}_Q J(Q_t, z_{t+1}; \mathbf{s}_t,\mathbf{a}_t,\mathbf{s}_t')\rangle_\ccalH 
 + \frac{L_Q \alpha_t^2}{2} \| \tilde{\nabla}_Q J(Q_t, z_{t+1}; \mathbf{s}_t,\mathbf{a}_t,\mathbf{s}_t') \|_\ccalH^2 \; .
\end{align}
Subsequently, we use the short-hand notation $\hat{\nabla}_Q J(Q_t):=\hat{\nabla}_Q J(Q_t, z_{t+1}; \mathbf{s}_t,\mathbf{a}_t,\mathbf{s}_t') $ and $\tilde{\nabla}_Q J(Q_t):=\tilde{\nabla}_Q J(Q_t, z_{t+1}; \mathbf{s}_t,\mathbf{a}_t,\mathbf{s}_t') $ for the stochastic and projected stochastic quasi-gradients, \eqref{eq:unprojectedgradproof} and \eqref{eq:projectedgradproof}, respectively. Now add and subtract the inner-product of the functional gradient of $J$ with the stochastic gradient, scaled by the step-size $ \alpha_t \langle \nabla J(Q_t) , \hat{\nabla}_Q J(Q_t, z_{t+1}; \mathbf{s}_t,\mathbf{a}_t,\mathbf{s}_t') \rangle_\ccalH $, and $\alpha_t \|\nabla_Q J(Q_t)\|^2$ into above expression and gather terms.
\begin{align} \label{eq:pbroof}
\!\!\!\!\!\!J(\!Q_{t+1}\!)\! &\! \leq   \! J(\!Q_t\!)\!-\! \alpha_t \|\!\nabla_{\!\!Q} J(Q)\|^2\! \!+\! \frac{L_Q \alpha_t^2}{2} \!\| \!\tilde{\nabla}_{\!\!Q} J(Q_t) \!\|_\ccalH^2 \!\!\!
 -\! \alpha_t\! \langle \nabla \!J(Q_t\!) , 
\!\!\tilde{\nabla\!\!}_Q J(\!Q_t\!) \!-\!\! \hat{\nabla}_Q J(Q_t\!) \rangle_\ccalH
+ \alpha_t   \langle \nabla J(Q_t) ,
\nabla J(Q_t)- \hat{\nabla}_Q J(Q_t) \rangle_\ccalH  \; 
\end{align}
Observe that the last two terms on the right-hand side of \eqref{eq:pbroof} are terms associated with the directional error between the true gradient and the stochastic quasi-gradient, as well as the stochastic quasi-gradient with respect to the projected stochastic quasi-gradient. The former term may be addressed through the error bound derived from the KOMP stopping criterion in Proposition \ref{prop_proj_grad_err}, whereas the later may be analyzed through  the Law of Total Expectation and Assumptions \ref{as:mean_variance_delta} - \ref{as:mean_variance_q}. 

We proceed to address the second term on the right-hand side of \eqref{eq:pbroof} by applying Cauchy-Schwarz to write
\begin{align} \label{eq:qproof}
|-\alpha_t\! & \langle \nabla \!J(Q_t\!) , 
\!\!\tilde{\nabla\!\!}_Q J(\!Q_t\!) \!-\!\! \hat{\nabla}_Q J(Q_t\!) \rangle_\ccalH| 
\leq \alpha_t \|  \nabla \!J(Q_t\!) \|_\ccalH  \| \tilde{\nabla\!\!}_Q J(\!Q_t\!) \!-\!\! \hat{\nabla}_Q J(Q_t\!) \|_\ccalH 
\end{align}
Now, apply Proposition \ref{prop_proj_grad_err} to $\| \tilde{\nabla\!\!}_Q J(\!Q_t\!) \!-\!\! \hat{\nabla}_Q J(Q_t\!) \|_\ccalH $, the Hilbert-norm error induced by sparse projections on the right-hand side of \eqref{eq:qproof} and cancel the factor of $\alpha_t$:
\begin{align} \label{eq:rproof}
\alpha_t\!  \langle \nabla \!J(Q_t\!) , 
\!\!\tilde{\nabla\!\!}_Q J(\!Q_t\!) \!-\!\! \hat{\nabla}_Q J(Q_t\!) \rangle_\ccalH  \leq \epsilon_t \|  \nabla \!J(Q_t\!) \|_\ccalH 
\end{align}
Next, we address the last term on the right-hand side of \eqref{eq:pbroof}. To do so, we will exploit Assumptions \ref{as:mean_variance_delta} - \ref{as:mean_variance_q} and the Law of Total Expectation. First, consider the expectation of this term, ignoring the multiplicative step-size factor, while applying \eqref{eq:finitevariance1proof}:
\begin{align}\label{eq:total_expectations_1}
& \mathbb{E}\big[ \nabla J(Q_t) ,
\nabla_Q J(Q_t)- \hat{\nabla}_Q J(Q_t\!) \rangle_\ccalH\given \ccalF_t \big] = \Big\langle\!\!\nabla_{\!\!\!Q} J(Q_t), \mathbb{E} \lbrack  (\!\gamma \kappa(\!(\!\mathbf{s}_t',\mathbf{a}_t'),\!\cdot\!)\! - \!\kappa(\!(\!\mathbf{s}_t,\mathbf{a}_t\!)\!,\!\cdot\!)\!)\!(\!\bar{\delta}_t  \!-\! z_{t+1}\! )\!\given\! \ccalF_t \rbrack \!\!\Big\rangle_{\!\!\!\ccalH}
\end{align}
In \eqref{eq:total_expectations_1}, we pull the expectation inside the inner-product, using the fact that $\nabla_Q J(Q)$ is deterministic. Note on the right-hand side of \eqref{eq:total_expectations_1}, by using \eqref{eq:finitevariance1proof}, we have $\bar{\delta}_t$ inside the expectation in the above expression rather than a realization $\delta_t$.
Now, apply Cauchy-Schwartz to the above expression to obtain
\begin{align}\label{eq:total_expectations_2}
&\Big\langle\!\!\nabla_{\!\!\!Q} J(Q_t), \mathbb{E} \lbrack  (\!\gamma \kappa(\!(\!\mathbf{s}_t',\mathbf{a}_t'),\!\cdot\!)\! - \!\kappa(\!(\!\mathbf{s}_t,\mathbf{a}_t\!)\!,\!\cdot\!)\!)\!(\!\bar{\delta}_t  \!-\! z_{t+1}\! )\!\given\! \ccalF_t \rbrack \!\!\Big\rangle_{\!\!\!\ccalH} \leq \|\!\nabla_{\!\!\!Q} J(Q_t\!)\!\|_{\ccalH} \mathbb{E} \lbrack  \|(\!\gamma \kappa(\!(\!\mathbf{s}_t',\mathbf{a}_t'),\!\cdot\!)\!\! - \!\kappa(\!(\!\mathbf{s}_t,\mathbf{a}_t\!)\!,\!\cdot\!)\!)\|_{\ccalH}  \times |\!\bar{\delta}_t  \!-\! z_{t+1}\! |\given\! \ccalF_t \rbrack 
\end{align}
From here, apply the inequality $ab \leq \frac{\rho}{2} a^2  + \frac{1}{2\rho} b^2$ for $\rho>0$ with  $a = |\bar{\delta_t} - z_{t+1}|$, and $b=\alpha_t \| \nabla_Q J(Q_t) \|  \| \gamma \kappa((\mathbf{s}_t',\mathbf{a}_t'),\cdot) - \kappa((\mathbf{s}_t,\mathbf{a}_t),\cdot)\|_\ccalH$, and $\rho=\beta_t$ to the preceding expression:
\begin{align}\label{eq:apply_tricky_inequality}
\!\!\|\!&\nabla_{\!\!\!Q} J(Q_t\!)\!\|_{\!\ccalH} \mathbb{E} \lbrack  \!\|\!(\!\gamma \kappa(\!(\!\mathbf{s}_t',\!\mathbf{a}_t')\!,\!\cdot\!)\!\! - \!\kappa(\!(\!\mathbf{s}_t,\!\mathbf{a}_t\!)\!,\!\cdot\!)\!)\!\|_{\ccalH}
\! |\!\bar{\delta}_t  \!-\! z_{t+1}\! |\!\! \given\!\! \ccalF_t \rbrack \!\!\Big\rangle_{\!\!\!\ccalH} \nonumber \\
&\leq \frac{\beta_t}{2} \mathbb{E} \lbrack ( \bar{\delta_t} - z_{t+1} )^2 \mid \mathcal{F}_t \rbrack  + \frac{\alpha_t^2}{2 \beta_t}\! \|\!\nabla J(Q_t)\! \|^2_\ccalH \! \mathbb{E}[\| \gamma \kappa(\!(\!\mathbf{s}_t',\mathbf{a}_t'),\cdot\!)\! -\! \kappa((\mathbf{s}_t,\mathbf{a}_t),\cdot)\|_\ccalH^2\given \ccalF_t]
\end{align}
To \eqref{eq:apply_tricky_inequality}, we apply Assumption \ref{as:mean_variance_q} regarding the finite second conditional of the difference of reproducing kernel maps \eqref{eq:finitevariance2proof} to the second term, which when substituted into the right-hand side of the expectation of \eqref{eq:pbroof} conditional on $\ccalF_t$, yields
\begin{align}\label{eq:lemma_iii_near_final}
\mathbb{E}[J(\!Q_{t+1}\!)\given\ccalF_t] &\leq   J(\!Q_t\!)
- \alpha_t \!\!\left(\!\!1\! -\! \frac{\alpha_t G_{Q}^2 }{ \beta_t}\!\!\right) \|\nabla_Q J(Q)\|^2\! \!
+\!  \frac{\beta_t}{2} \mathbb{E} \lbrack ( \bar{\delta_t} - z_{t+1} )^2 \mid \mathcal{F}_t \rbrack+\frac{L_Q \sigma_Q^2 \alpha_t^2}{2}  + \eps_t \|\nabla_Q J(Q_t) \|_{\ccalH} \; ,
\end{align}
where we have also applied the fact that the projected stochastic quasi-gradient has finite conditional variance \eqref{eq:finitemomentproof} and gathered like terms to conclude \eqref{eq:lemma1iiiproof}. 
\end{myproof}

Lemma \ref{main_lemma} is may be seen as a nonparametric extension of Lemma 2 and A.1 of \cite{wang2017stochastic}, or an extension of Lemma 6 in \cite{pkgtd} to the non-convex case. Now, we may use Lemma \ref{main_lemma} to connect the function sequence generated by Algorithm \ref{qlearning} to a special type of stochastic process called a coupled supermartingale, and therefore prove that $Q_t$ converges to a stationary point of the Bellman error with probability $1$. To the best of our knowledge, this is a one of a kind result.
\subsection*{Appendix B: Proof of Theorem \ref{main_theorem}}\label{sec:apx_b}

We use the relations established in Lemma \ref{main_lemma} to construct a coupled supermatingale of the form \ref{lemma_coupled_supermartingale}. First, we state the following lemma regarding coupled sequences of conditionally decreasing stochastic processes called the coupled supermartingale lemma, stated as:

\begin{lemma}\label{lemma_coupled_supermartingale}
(Coupled Supermartingale Theorem) \cite{bertsekas2004stochastic,wang2017stochastic}. Let $\lbrace \xi_tt\rbrace$, $\lbrace \zeta_t \rbrace$, $\lbrace u_t \rbrace$, $\lbrace \bar{u}_t \rbrace$, $\lbrace \eta_t \rbrace$, $\lbrace \theta_t \rbrace$, $\lbrace \epsilon_t \rbrace$, $\lbrace \mu_t \rbrace$, $\lbrace \nu_t \rbrace$ be sequences of  nonnegative random variables such that
\begin{align} \label{eq:zbproof}
&\mathbb{E} \lbrace \xi_{t+1} \mid G_t \rbrace \leq (1+\eta_t)\xi_t - u_t + c\theta_t\zeta_t + \mu_t \;,
\nonumber \\ 
&\mathbb{E} \lbrace \zeta_{t+1} \mid G_t \rbrace \leq (1-\theta_t)\zeta_t - \bar{u}_t + \epsilon_t\xi_t + \nu_t
\end{align}
where $G_t = \lbrace \xi_s, \xi_s, u_s,\bar{u}_s, \eta_s, \theta_s, \epsilon_s, \mu_s, \nu_s \rbrace_{s=0}^t$ is the filtration  and $c > 0$ is a scalar. Suppose the following summability conditions hold almost surely:
\begin{align} \label{eq:zcproof}
\sum_{t=0}^\infty \eta_t < \infty
\;,
\sum_{t=0}^\infty \epsilon_t < \infty
\;,
\sum_{t=0}^\infty \mu_t < \infty
\;,
\sum_{t=0}^\infty \nu_t < \infty
\end{align}
Then $\xi_t$ and $\zeta_t$ converge almost surely to two respective nonnegative random variables, and we may conclude that almost surely
\begin{equation} \label{eq:zdproof}
\sum_{t=0}^\infty u_t < \infty
\;,
\sum_{t=0}^\infty \bar{u}_t < \infty
\;,
\sum_{t=0}^\infty \theta_t \zeta_t < \infty
\end{equation}
\end{lemma}

We construct coupled supermartingales that match the form of Lemma \ref{lemma_coupled_supermartingale} using Lemma \ref{main_lemma}. First, use Lemma \ref{main_lemma}\eqref{main_lemma_2}\eqref{eq:lemma1iiproof} as an upper bound on Lemma \ref{main_lemma}\eqref{main_lemma_3} \eqref{eq:lemma1iiiproof}.
\begin{align} \label{eq:zeproof}
\mathbb{E} [J  (\!Q_{t+1}\!) \given  \ccalF_t] \leq & J(\!Q_t\!)
- \alpha_t \!\!\left(\!\!1\! -\! \frac{\alpha_t G_{Q}^2 }{ \beta_t}\!\!\right) \|\nabla_Q J(Q_t)\|^2\! \!
\nonumber \\ 
& + \!  \frac{\beta_t}{2} ((1-\beta_t)(z_t - \bar{\delta}_{t-1})^2   + \frac{L_Q}{\beta_t} \|Q_t - Q_{t-1} \|_\ccalH^2 
+ \! 2 \beta_t^2 \sigma_\delta^2) \!+ \! \frac{L_Q \sigma_Q^2 \alpha_t^2}{2} + \eps_t \|\!\nabla_{\!Q} J(Q_t) \|_{\ccalH} 
\end{align}
We introduce three restrictions on the learning rate, expectation rate, and parsimony constant in order to simplify \eqref{eq:zeproof}. First, we assume that $\beta_t \in (0,1)$ for all $t$. Next, we choose $\epsilon_t = \alpha_t^2$. Lastly, we restrict $1 - \frac{\alpha_t G_{Q}^2 }{ \beta_t}  > 0 $, which results in the condition: $\frac{\alpha_t  }{ \beta_t} < \frac{1}{G_{Q}^2}$. Then, we simplify and group terms of \eqref{eq:zeproof}.
\begin{align} \label{eq:zfproof}
\mathbb{E}[J(\!Q_{t+1}\!)\given  \ccalF_t] \leq &   J(\!Q_t\!)
- \alpha_t \|\nabla_Q J(Q_t)\|^2\! \!
+ \!  \frac{\beta_t}{2}(z_t - \bar{\delta}_{t-1})^2   + \frac{L_Q}{2} \|Q_t - Q_{t-1} \|_\ccalH^2 
+ \beta^3_t \sigma_\delta^2 + \alpha_t^2 \left( \!\!\! \frac{L_Q \sigma_Q^2 }{2} +  \|\nabla_Q J(Q_t) \|_{\ccalH}\! \!\!\!\right) \;
\end{align}
Next, we aim to connect the result of \eqref{eq:zeproof} to the form of Lemma \ref{lemma_coupled_supermartingale} via the identifications:
\begin{align} \label{eq:zgproof}
&\xi_t  = J(Q_{t}) \;,  
\zeta_t  = (z_t - \bar{\delta}_{t-1})^2 \;, \theta_t = \beta_t\;, c = 1/2
\\ \nonumber
& u_t = \alpha_t \|\nabla_Q J(Q_t)\|^2\! \! \;, \eta_t  = 0  
\\ \nonumber
& \mu_t \! = \!\! \frac{L_Q}{2} \|Q_t \!\! - \! Q_{t\!-\!1} \|_\ccalH^2 \!\! + \! \beta^3_t \! \sigma_\delta^2 \! + \! \alpha_t^2\! \!\!\left( \!\!\! \frac{L_Q \sigma_Q^2 }{2} \!\!+\!  \|\nabla_{\!Q} J(Q_t) \|_{\ccalH}\!\!\! \right) 
\end{align}
Observe that $\sum \mu_t < \infty$ due to the upper bound on $\|Q_t \!\! - \! Q_{t\!-\!1} \|_\ccalH^2$ provided by Lemma \ref{main_lemma}\eqref{eq:lemma1iproof} and the summability conditions for $\alpha_t^2$ and $\beta_t^2$ \eqref{eq:theorem1proof}.  

Next, we identify terms in Lemma  \ref{main_lemma} \eqref{main_lemma_2} \eqref{eq:lemma1iiproof} according to Lemma \ref{lemma_coupled_supermartingale} in addition to \eqref{eq:zgproof}.
\begin{align} \label{eq:zhproof}
\nu_t = \frac{L_Q}{\beta_t} \|Q_t \! - \! Q_{t-1} \|_\ccalH^2 \! + \! 2 \beta^2_t \sigma_\delta^2  \;, \epsilon_t = 0 \;, \bar{u}_t = 0
\end{align}
The summability of $\nu_t$ can be shown as follows: the expression $\| Q_t - Q_{t-1}\|^2_\ccalH / \beta_t$ which is order $\mathcal{O}(\alpha_t^2/\beta_t)$ in conditional expectation by Lemma  \ref{main_lemma}\eqref{main_lemma_1}. Sum the resulting conditional expectation for all $t$, which by the summability of the sequence $\sum_t \alpha_t^2 / \beta_t < \infty$ is finite. Therefore, $\sum_t \| Q_t - Q_{t-1}\|^2_\ccalH / \beta_t < \infty$ almost surely.  We also require $\sum_t \beta_t^2 < \infty$ \eqref{eq:theorem1proof} for the summability of the second term of \eqref{eq:zhproof}

Together with the conditions on the step-size sequences $\alpha_t$ and $\beta_t$, the summability conditions of Lemma \ref{lemma_coupled_supermartingale} are satisfied, which allows to conclude that $\xi_t = J(Q_{t}) $ and $\zeta_t = (z_t - \bar{\delta}_{t-1})^2$ converge to two nonnegative random variables w.p. 1, and that
\begin{align} \label{eq:zjproof}
\!\!\sum_t \!\alpha_t \|\nabla_Q J(Q_t)\|^2 < \infty ,\quad  \sum_t \beta_t (z_t - \bar{\delta}_{t-1})^2 < \infty \;
\end{align}
almost surely. Then, the summability of $u_t$ taken together with non-summability of $\alpha_t$ and $\beta_t$ \eqref{eq:theorem1proof} indicates that the limit infimum of the norm of the gradient of the cost goes to null. 
\begin{align} \label{eq:zkproof}
\liminf_{t \rightarrow \infty} \|\nabla_Q J(Q_t) \| = 0 \;, 
\liminf_{t \rightarrow \infty} (z_t - \bar{\delta}_{t-1})^2 = 0 
\end{align}
almost surely. 
From here, given $\liminf_{t \rightarrow \infty} \|\nabla_Q J(Q_t) \|_{\ccalH} = 0$, we can apply almost the exact same argument by contradiction as \cite{wang2017stochastic} to conclude that the whole sequence $\|\nabla_Q J(Q_t) \|_{\ccalH}$ converges to null with probability $1$, which is repeated here for completeness.

 Consider some $\eta>0$ and observe that $\|\nabla_Q J(Q_t)\|_{\ccalH}\leq \eta$ for infinitely many $t$. Otherwise, there exists $t_0$ such that $\sum_{t=0}^\infty \|\alpha_t\nabla_Q  J(Q_t)\|_{\ccalH}^2 \geq \sum_{t=t_0}^\infty \alpha_t \eta^2 =\infty$ which contradicts \eqref{eq:zjproof}. Therefore, there exists a closed set $\bar{\ccalH} \subset \ccalH$ such that $\{Q_t\}$ visits $\bar{\ccalH}$ infinitely often, and
 \begin{align}\label{eq:cases_functional_gradient1}
 \|\nabla_Q J(Q) \|_{\ccalH} 
 \begin{cases} 
 \leq \eta \text{ for } Q\in\bar{\ccalH} \\
 > \eta \text{ for  } Q \not\in\bar{\ccalH}, Q \in \{Q_t \}
 \end{cases}
 \end{align}
Suppose to the contrary that there exists a limit point $\tilde{Q}$ such that $\|\nabla_Q J(\tilde{Q})\|_{\ccalH} > 2 \eta$. Then there exists a closed set $\tilde{\ccalH}$, i.e., a union of neighborhoods of all $Q_t$'s {such that $\|\nabla_Q J(Q_t)\|_{\ccalH} > 2 \eta$, with $\{Q_t\}$ visiting $\tilde{\ccalH}$ infinitely often}, and 
 \begin{align}\label{eq:cases_functional_gradient2}
 \|\nabla_Q J(Q) \|_{\ccalH} 
 \begin{cases} 
 \geq 2\eta \text{ for } Q\in\tilde{\ccalH} \\
 < 2\eta \text{ for  } Q \not\in\tilde{\ccalH}, Q \in \{Q_t \}
 \end{cases}
 \end{align}
Using the continuity of $\nabla J$ and $\eta>0$, we have that $\bar{\ccalH}$ and $\tilde{\ccalH}$ are disjoint: $\text{dist}(\bar{\ccalH},\tilde{\ccalH}) > 0$. Since $\{Q_t\}$ enters both $\bar{\ccalH}$ and $\tilde{\ccalH}$ infinitely often, there exists a subsequence {$\{Q_t\}_{t\in\ccalT} = \{ \{Q_t\}_{t=k_i}^{j_i-1}\}$} (with $\ccalT \subset \mathbb{Z}^+$) that enters  $\bar{\ccalH}$ and $\tilde{\ccalH}$ infinitely often, with $Q_{k_i} \in  \bar{\ccalH}$ and $Q_{j_i}\in \tilde{\ccalH}$ for all $i$. Therefore, for all $i$, we have
\begin{align}\label{eq:grad_inequality_eta}
&\|\nabla_Q J(Q_{k_i}) \|_{\ccalH} \geq 2 \eta> \|\nabla_Q J(Q_{t}) \|_{\ccalH}    \\
&> \eta \geq \|\nabla_Q J(Q_{j_i}) \|_{\ccalH} \quad\text{ for } t = k_i +1 , \dots, j_i -1\nonumber
\end{align}
Therefore, we can write 
\begin{align}\label{eq:divergence_setdist}
\sum_{t\in\ccalT} \|Q_{t+1} - Q_t \|_{\ccalH} &=\sum_{i=1}^\infty \sum_{t={k_i}}^{j_i-1} \|Q_{t+1} - Q_t \|_{\ccalH}  \geq \sum_{i=1}^\infty \|Q_{k_i} - Q_{j_i} \|_{\ccalH} \geq \text{dist}(\bar{\ccalH},\tilde{\ccalH}) = \infty
\end{align}
However, we may also write that 
\begin{align}\label{eq:convergence_gradient}
\infty> \sum_{t=0}^\infty \alpha_t \|\nabla J(Q_t) \|_{\ccalH}^2 \geq \sum_{t\in\ccalT} \alpha_t \|\nabla J(Q_t) \|_{\ccalH}^2 > \eta^2 \sum_{t\in\ccalT} \alpha_t 
\end{align}
Then, using the fact that the sets $\ccalX$ and $\ccalA$ are compact, there exist some $M>0$ such that \\ $\|Q_{t+1} - Q_t \|_{\ccalH}\leq \alpha_t \| \tilde{\nabla}_Q J(Q_t, z_{t+1}; \mathbf{s}_t,\mathbf{a}_t,\mathbf{s}_t')\|_{\ccalH} \leq M \alpha_t$ for all $t$, using the fact that $\eps_t = \alpha_t^2$. Therefore, 
\begin{align}\label{eq:divergence_setdist2}
\sum_{t\in\ccalT} \|Q_{t+1} - Q_t \|_{\ccalH} \leq M \sum_{t\in\ccalT} \alpha_t <\infty
\end{align}
which contradicts \eqref{eq:divergence_setdist}. Therefore, there does not exist any limit point $\tilde{Q}$ such that $\|\nabla_Q J(\tilde{Q}) \|_{\ccalH} > 2 \eta$. By making $\eta$ arbitrarily small, it means that there does not exist any limit point that is nonstationary. Moreover, we note that the set of such sample paths occurs with probability $1$, since the preceding analysis applies to all sample paths which satisfy \eqref{eq:zjproof}. Thus, any limit point of $Q_t$ is a stationary point of $J(Q)$ almost surely. $\hfill \blacksquare$
%

\subsection*{Appendix C: Proof of Theorem \ref{main_theorem2}}\label{apx_theorem2}
%
Begin with the expression in \eqref{eq:lemma1iiiproof}, and substitute in constant step-size selections with $\eps=C \alpha^2$ and $(1-\beta) \leq 1$, and compute the total expectation ($\ccalF_t=\ccalF_0$ ) to write
\begin{align} \label{eq:lemma1iiiproof_constant}
\mathbb{E}[J(\!Q_{t+1}\!)] &\leq   \mathbb{E}[J(\!Q_t\!)]
- \alpha \!\!\left(\!\!1\! -\! \frac{\alpha G_{Q}^2 }{ \beta}\!\!\right) \mathbb{E}[\|\nabla_Q J(Q)\|^2 ]
+ C \alpha^2 \mathbb{E}[\|\nabla_Q J(Q_t) \|_{\ccalH} ] +\!  \frac{\beta}{2} \mathbb{E} \lbrack (z_{t+1}-  \bar{\delta_t} )^2 \rbrack +\frac{L_Q \sigma_Q^2 \alpha^2}{2} \; ,%
\end{align}
From here, we note that the sequence $\mathbb{E} \lbrack (z_{t+1}-  \bar{\delta_t} )^2 \rbrack$ is identical (except re-written in terms of Q-functions rather than value functions) to the sequence in Lemma 1(iii) of \cite{pkgtd}, and therefore, analogous reasoning regarding to that which yields eqn. (86) in Appendix D of \cite{pkgtd} allows us to write
\begin{align} \label{eq:lemma1iiproof_constant}
\!\!\!\!\!\!\mathbb{E} \lbrack  (z_{t+1}\! -  \!\bar{\delta_t})^2  \rbrack & \leq  \frac{2 L_Q}{\beta^2}\left[\alpha^2 (G_\delta^2 G_Q^2\! +\! \lambda^2 D^2\right] + 2 \beta \sigma_\delta^2 \; ,
\end{align}
which follows from applying \eqref{eq:lemma1iproof} to the Hilbert-norm difference of $Q$-functions term and recursively substituting the total expectation of \eqref{eq:lemma1iiproof} back into itself, and simplifying the resulting geometric sum. Now, we may substitute the right-hand side of \eqref{eq:lemma1iiproof_constant} into the third term on the right-hand side of \eqref{eq:lemma1iiiproof_constant} to obtain
\begin{align} \label{eq:lemma1iiiproof_constant2}
\mathbb{E}[J&(\!Q_{t+1}\!)] \leq   \mathbb{E}[J(\!Q_t\!)]
- \alpha \!\!\left(\!\!1\! -\! \frac{\alpha G_{Q}^2 }{ \beta}\!\!\right) \mathbb{E}[\|\nabla_Q J(Q)\|^2 ]
+ C \alpha^2 \mathbb{E}[\|\nabla_Q J(Q_t) \|_{\ccalH} ] +\!     \frac{2 L_Q}{\beta}\left[\alpha^2 (G_\delta^2 G_Q^2\! +\! \lambda^2 D^2\right]  + 2 \beta^2 \sigma_\delta^2 +\frac{L_Q \sigma_Q^2 \alpha^2}{2} \; ,%
\end{align}
The rest of the proof proceeds as follows: we break the right-hand side of \eqref{eq:lemma1iiiproof_constant2} into two subsequences, one in which the expected norm of the cost functional's gradient  $\mathbb{E}[\|\nabla_Q J(Q_t) \|_{\ccalH}]$ is below a specified threshold, whereby $J(Q_t)$ is a decreasing sequence in expectation, and one where this condition is violated. We can use this threshold condition to define a deterministically decreasing auxiliary sequence to which the Monotone Convergence Theorem applies, and hence we obtain convergence of the auxiliary sequence. Consequently, we obtain convergence in infimum of the expected value of $J(Q_t)$ to a neighborhood.

We proceed by defining the conditions under which $\mathbb{E}[J(Q_{t})]$ is decreasing, i.e.,
\begin{align} \label{eq:objective_decreasing_condition}
\mathbb{E}[J&(\!Q_{t+1}\!)] \leq   \mathbb{E}[J(\!Q_t\!)]
- \alpha \!\!\left(\!\!1\! -\! \frac{\alpha G_{Q}^2 }{ \beta}\!\!\right) \mathbb{E}[\|\nabla_Q J(Q)\|^2 ]
+ C \alpha^2 \mathbb{E}[\|\nabla_Q J(Q_t) \|_{\ccalH} ] +\!     \frac{2 L_Q}{\beta}\left[\alpha^2 (G_\delta^2 G_Q^2\! +\! \lambda^2 D^2\right] + 2 \beta^2 \sigma_\delta^2 +\frac{L_Q \sigma_Q^2 \alpha^2}{2}  \nonumber \\
&\qquad\quad \leq  \mathbb{E}[J(\!Q_t\!)]
\end{align}
Note that \eqref{eq:objective_decreasing_condition} holds whenever the following is true:
\begin{align} \label{eq:gradient_condition}
&\hspace{-4mm}- \alpha \!\!\left(\!\!\!1\! -\! \frac{\alpha G_{Q}^2 }{ \beta}\!\!\!\right) \!\!\mathbb{E}[\|\nabla_Q J(Q)\|^2 ]
+ C \alpha^2 \mathbb{E}[\|\nabla_Q J(Q_t) \|_{\ccalH} ]      \frac{2 L_Q}{\beta}\left[\alpha^2 (G_\delta^2 G_Q^2\! +\! \lambda^2 D^2\right] 
+ 2 \beta^2 \sigma_\delta^2 +\frac{L_Q \sigma_Q^2 \alpha^2}{2}   \leq 0
\end{align}
Observe that the left-hand side of \eqref{eq:gradient_condition} defines a quadratic function of $\mathbb{E}[\|\nabla_Q J(Q_t) \|_{\ccalH} ]$ which opens downward. We can solve for the condition under which \eqref{eq:gradient_condition} holds with equality by obtaining the positive root (since $\mathbb{E}[\|\nabla_Q J(Q_t) \|_{\ccalH} ] \geq 0$) of this expression through the quadratic formula:
\begin{align} \label{eq:quadratic_formula}
\mathbb{E}[\|\!\nabla_{\!\!Q} J(Q)\| ] \! &= \!\!\Bigg(\!\!\!C  \!\!+\!\! \Bigg[\!C^2  \!\! +\! 4  \!\!\left(\!\!\!\!\frac{1}{\alpha}\! -\! \frac{ G_{Q}^2 }{ \beta}\!\!\!\right)\!\!\!\Big[ \!\frac{2 L_Q}{\beta}\!\!\left[ \!(\!G_\delta^2 G_Q^2\! +\! \lambda^{\!\!\!2} D^2)\!\right]  
 +\! \frac{2 \beta^2 \sigma_\delta^2}{\alpha^2}  \!+\!\frac{L_Q \sigma_Q^2 }{2}\Big]\!\! \Bigg]^{\!\!\!1/2}\Bigg)\!\!\!\left(\!\!\!\!\frac{1}{\alpha} \!-\! \frac{ G_{Q}^2 }{ \beta}\!\!\!\right)^{-1} \hspace{-2mm}  \\
&= \ccalO\!\!\left( \!\!\frac{\alpha \beta}{\beta - \alpha} \!\! \left[1 + \sqrt{1 + \frac{\beta - \alpha}{\alpha \beta}\left(\frac{1}{\beta} + \frac{\beta^2}{\alpha^2}  \right)} \!\!\! \right]\!\!   \right)\nonumber
\end{align} 
where we have cancelled out a common factor of $\alpha^2$ as well as common factors of $-1$ throughout. Now, define the right-hand side of \eqref{eq:quadratic_formula} as a constant $\Delta$, and the auxiliary sequence 
\begin{align} \label{eq:aux_sequence}
\Gamma_t= \mathbb{E}[J(Q_t)]\mathbbm{1}\Big\{& \min_{u \leq t} - \alpha \!\!\left(\!\!\!1\! -\! \frac{\alpha G_{Q}^2 }{ \beta}\!\!\!\right) \!\!\mathbb{E}[\|\nabla_Q J(Q)\|^2 ]  + C \alpha^2 \mathbb{E}[\|\nabla_Q J(Q_t) \|_{\ccalH} ]+\!     \frac{2 L_Q}{\beta}\big[\alpha^2 (G_\delta^2 G_Q^2\! 
+\! \lambda^2 D^2\big] + 2 \beta^2 \sigma_\delta^2 +\frac{L_Q \sigma_Q^2 \alpha^2}{2}   > \Delta  \Big\}
\end{align}
where $\mathbbm{1}\{E\}$ denotes the indicator function of a (deterministic) event $E$. From here, we note that $\Gamma_t$ is nonnegative since $J(Q_t) \geq 0$. Moreover,  $\Gamma_t$ is decreasing: either the indicator is positive, in which case its argument is true, and hence \eqref{eq:gradient_condition} holds with equality. When \eqref{eq:gradient_condition} holds with equality, the objective is decreasing, namely, \eqref{eq:objective_decreasing_condition} is valid. Alternatively, condition inside the indicator is null, which, due to the use of the minimum in the definition \eqref{eq:aux_sequence}, means that the indicator is null for all subsequent times. Therefore, in either case, $\Gamma_t$ is nonnegative and decreasing, and therefore we may apply the Monotone Convergence Theorem \cite{rudin-principles} to conclude $\Gamma_t \rightarrow 0$. Therefore, we have either that $\lim_t \mathbb{E}[J(Q_t)] - \Delta = 0$ or that the indicator in \eqref{eq:aux_sequence} is null for $t\rightarrow \infty$. Taken together, these statements allow us to conclude
\begin{align}\label{eq:conclusion_constant_stepsize}
\liminf_{t\rightarrow \infty} \ \mathbb{E}[J(Q_t)] &\leq \ccalO\!\!\left(\! \!\!\frac{\alpha \beta}{\beta - \alpha} \!\!\! \left[\!\!1 \!+\! \sqrt{\!\!1\! +\! \frac{\beta - \alpha}{\alpha \beta}\!\!\!\left(\!\frac{1}{\beta} \!+ \!\frac{\beta^2}{\alpha^2}  \!\!\right)} \!\!\! \right]\!\!   \right)
\end{align}
which is as stated in \eqref{eq:theorem2}.$\hfill \blacksquare$

%

\end{document}